\documentclass[10pt,twocolumn,letterpaper]{article}

\usepackage[pagenumbers]{cvpr} 

\definecolor{cvprblue}{rgb}{0.21,0.49,0.74}
\usepackage[pagebackref,breaklinks,colorlinks,allcolors=cvprblue]{hyperref}

\usepackage{natbib}
\usepackage{float}
\usepackage{multirow}
\usepackage{makecell}
\usepackage{graphicx}

\usepackage{algorithm}
\usepackage{listings}
\usepackage{amsmath}
\usepackage{amsfonts}
\usepackage{tikz}

\usepackage{pifont}
\usepackage{color}
\usepackage{colortbl}
\usepackage{xspace}
\definecolor{bluee}{RGB}{225, 235, 246}
\definecolor{upup}{RGB}{255,0,0}
\definecolor{down}{RGB}{83,100,147}

\def\paperID{14233} 
\def\confName{CVPR}
\def\confYear{2025}

\title{MiniMaxAD: A Lightweight Autoencoder for Feature-Rich Anomaly Detection}

\author{%
Fengjie Wang~~~~Chengming Liu~~~~Lei Shi~~~~Pang Haibo$\thanks{Corresponding Author}$ \\
Zhengzhou University, China\\
\texttt{wangfj515@foxmail.com} \quad
\texttt{cmliu@zzu.edu.cn} \\
\texttt{shilei@zzu.edu.cn} \quad
\texttt{panghbzzu@163.com}
}
\begin{document}
\maketitle
\begin{abstract}
  Previous industrial anomaly detection methods often struggle to handle the extensive diversity in training sets, particularly when they contain stylistically diverse and feature-rich samples, which we categorize as feature-rich anomaly detection datasets (FRADs). This challenge is evident in applications such as multi-view and multi-class scenarios. To address this challenge, we developed MiniMaxAD, a efficient autoencoder designed to efficiently compress and memorize extensive information from normal images. Our model employs a technique that enhances feature diversity, thereby increasing the effective capacity of the network. It also utilizes large kernel convolution to extract highly abstract patterns, which contribute to efficient and compact feature embedding. Moreover, we introduce an Adaptive Contraction Hard Mining Loss (ADCLoss), specifically tailored to FRADs. In our methodology, any dataset can be unified under the framework of feature-rich anomaly detection, in a way that the benefits far outweigh the drawbacks. Our approach has achieved state-of-the-art performance in multiple challenging benchmarks. Code is available at: \href{https://github.com/WangFengJiee/MiniMaxAD}{https://github.com/WangFengJiee/MiniMaxAD}
\end{abstract}

\section{Introduction}
Anomaly detection plays a crucial role in quality control within modern intelligent manufacturing and has attracted growing attention in recent years.
Due to the challenges in collecting and labeling anomaly samples, a growing number of researchers are reducing their reliance on supervisory signals \cite{bozicMixedSupervisionSurfacedefect2021}. As a result, unsupervised learning methods are becoming increasingly popular for anomaly detection.
These methods utilize only normal samples for training and are engineered to identify anomalous images and segment anomalous regions during inference.
Recently, unsupervised anomaly detection (UAD) technology is increasingly used in more complex scenarios to meet growing demand.

Previous anomaly detection methods often require training a separate model for each class of objects, which can lead to significant memory consumption. In response, UniAD \cite{youUnifiedModelMulticlass2022} proposes a unified setting that trains a single model on multiple classes of normal samples. They also highlight the limitations of the traditional one-category-one-model approach, especially in cases where normal samples exhibit considerable intra-class diversity. This challenge is especially pronounced in the fields of unmanned supermarkets and multi-view anomaly detection, as illustrated by the newly introduced GoodsAD \cite{zhangPKUGoodsADSupermarketGoods2024} and Real-IAD \cite{wangRealIADRealworldMultiview2024} datasets, which exhibit significant intra-class diversity.

\begin{figure*}[t]
  \centering
  \includegraphics[width=\textwidth]{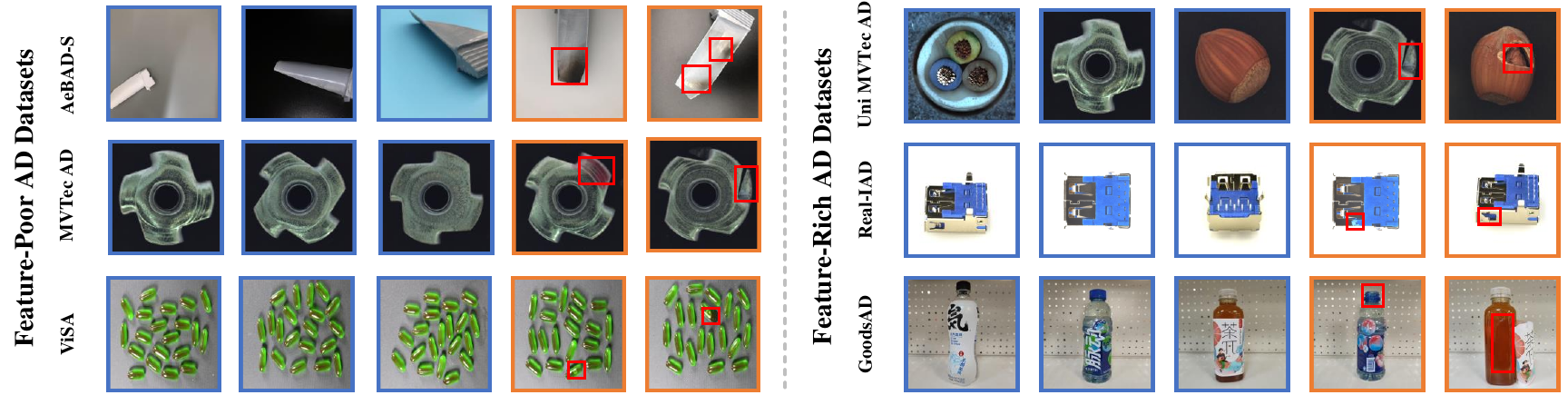}
  \caption{Examples of Feature-Rich Anomaly Detection Datasets (FRADs) and Feature-Poor Anomaly Detection Datasets (FPADs). Uni MVTecAD, i.e. MVTecAD under unified settings, represents a scenario where multiple categories share a model. Anomalous areas are marked with red boxes.}
  \label{fig1}
\end{figure*}

In this study, we revisited the existing UAD dataset and categorized it based on the diversity of samples within the training set, a concept we will refer to as \textbf{intra-set} diversity.
We define datasets with substantial intra-set diversity as Feature-Rich Anomaly Detection Datasets (FRADs), in contrast to Feature-Poor Anomaly Detection Datasets (FPADs), as illustrated in Figure~\ref{fig1}. Notably, FPADs can be easily converted to FRADs using the multi-class setting.
This diversity necessitates that neural networks extensively memorize normal features. Among FRADs, PatchCore \cite{rothTotalRecallIndustrial2022} demonstrates superior performance in applications such as unmanned supermarkets \cite{zhangPKUGoodsADSupermarketGoods2024} and multi-view scenarios \cite{wangRealIADRealworldMultiview2024}, benefiting from an efficient memory bank that stores normal contexts. However, this approach also significantly increases inference times due to its reliance on time-consuming query operations. Additionally, as the number of training samples grows, so does the memory bank's size, complicating the application of PatchCore in unified settings. Consequently, we aim to develop a high-performance, effective UAD method that addresses the challenges of FRADs effectively.

Methods based on multi-scale reconstruction \cite{dengAnomalyDetectionReverse2022,guoReContrastDomainspecificAnomaly2023,tienRevisitingReverseDistillation2023,zhangIndustrialAnomalyDetection2023} are celebrated for their simplicity and efficiency. These methods use a paired encoder and decoder to reconstruct features, effectively capturing the manifold of normal samples. During testing, areas with higher reconstruction errors are flagged as anomalies. However, they face two primary challenges: (a) the network's limited information capacity may inadequately reconstruct normal areas, resulting in false positives; (b) its robust generalization ability might allow it to accurately reconstruct abnormal areas, leading to false negatives, a phenomenon often referred to as the ``identical shortcut'' \cite{youUnifiedModelMulticlass2022}. Experimental results from GoodsAD \cite{zhangPKUGoodsADSupermarketGoods2024} show that these methods tend to generate a high rate of false positives by mistakenly classifying many normal foreground regions as abnormal in FRADs. Moreover, replacing the backbone of Reverse Distillation (RD) \cite{dengAnomalyDetectionReverse2022} with a larger model, such as WideResNet-101 \cite{zagoruykoWideResidualNetworks2017}, significantly reduces performance, as illustrated in Figure~\ref{fig3}(a). This suggests that dual challenges of limited capacity and the ``identical shortcut'' are present when dealing with FRADs.

In response to the dual challenges, our study introduces a core motivation: \textbf{maximizing the ``effective capacity'' of a relatively lightweight autoencoder}. Our approach is built on an efficient and compact architecture, the Reverse Distillation (RD) Architecture. However, it suffers from severe performance degradation when confronted with FRADs, particularly on the GoodsAD dataset, where the decline is critical—it nearly fails to reconstruct any foreground regions of the objects. To address this, we adopted a more aggressive improvement strategy by starting from its fundamental architecture. Our initial step is to reduce the number of parameters, which offers the dual benefits of enhancing computational efficiency and mitigating the ``identical shortcut'' issue. Although reducing the number of parameters typically results in decreased information capacity due to fewer neurons, we recognize that the ratio of information capacity to the number of parameters may have been underestimated. Despite having fewer parameters, a significant number of features can still be retained. To achieve this, we utilize large kernel convolutions to extract more comprehensive representations with relatively fewer parameters. Additionally, we employ a technique to prevent neuron inactivation, GRN \cite{wooConvNeXtV2CoDesigning2023}, which enhances network capacity by reactivating saturated or inactive neurons, thereby enabling them to represent more information. Finally, we introduce a novel Adaptive Contraction Hard Mining Loss (ADCLoss). Unlike previous hard mining methods, our approach allows the mining rate to adapt dynamically based on the characteristics of the given dataset. Our main contributions are summarized below:

\begin{itemize}
  \item We revisit existing UAD datasets and categorize them into feature-rich and feature-poor anomaly detection datasets based on intra-set diversity. The goal is to develop more targeted solutions for specific scenarios and to provide a framework for unifying these diverse datasets.

  \item We study the performance degradation of reconstruction-based methods in the presence of significant intra-set diversity from the perspective of information capacity and propose an effective solution.

  \item We developed a novel ADCLoss tailored for FRADs with excellent performance.

  \item Our comprehensive experiments show that our method achieves state-of-the-art performance across multiple challenging tasks.
\end{itemize}
\section{Related Work}
\label{sec:related}
\subsection{Unsupervised Anomaly Detection}
\textbf{Methods based on reconstruction.} In unsupervised anomaly detection (UAD), methods based on image-level reconstruction train models to reconstruct normal images, leveraging the assumption that anomalies will result in poor reconstruction quality due to model unfamiliarity with normal images. Recently, OCR-GAN \cite{liangOmniFrequencyChannelSelectionRepresentations2023} improved the performances by decoupling input images into various frequency components, employing multiple generators for enhanced reconstruction. TDAD \cite{weiTDADSelfsupervisedIndustrial2025} employs advanced diffusion models to achieve high-quality reconstruction. On the feature level, the introduction of the teacher-student model \cite{bergmannUninformedStudentsStudentteacher2020,guRememberingNormalityMemoryguided2023,dengAnomalyDetectionReverse2022, batznerEfficientADAccurateVisual2023, guoReContrastDomainspecificAnomaly2023}, inspired by knowledge distillation \cite{hintonDistillingKnowledgeNeural2015}, utilizes a pre-trained network as a 'teacher' to guide a 'student' model trained only on normal images. This approach targets feature reconstruction. Recently, MambaAD \cite{heMambaADExploringState2024}, based on reconstruction, explored the application of Mamba \cite{guMambaLinearTimeSequence2024} to unsupervised anomaly detection (UAD).

\textbf{Methods based on simulated anomalies.} This approach, which simulates anomalies to provide supervisory signals, includes various innovative methods \cite{tienRevisitingReverseDistillation2023,liuSimpleNetSimpleNetwork2023,zhangDeSTSegSegmentationGuided2023, yangSLSGIndustrialImage2023,songAnoSegAnomalySegmentation2021}. For instance, \cite{liCutPasteSelfsupervisedLearning2021b} utilizes a technique where a portion of an image is cut and randomly pasted elsewhere to create forged anomalies. Moreover, DRAEM \cite{zavrtanikDRAEMDiscriminativelyTrained2021} employs Perlin noise to construct anomalies and segments these during training to generalize the segmentation capability to actual anomalies. More recently, MemSeg \cite{yangMemSegSemisupervisedMethod2023} attempted to simulate anomalies within the foreground region of images. However, the segmentation of the foreground region proved to be unstable. Furthermore, SimpleNet \cite{liuSimpleNetSimpleNetwork2023} introduces gaussian noise into the feature space to simulate anomalies. The effectiveness of these methods hinges on the extent to which the pseudo anomalies resemble actual anomalies and on the placement of these fabricated anomalies. Generalizing these methods to new datasets remains challenging, and there is considerable scope for improvement.

\textbf{Methods Based on K-Nearest Neighbor (KNN).} KNN-based methods leverage feature matching for inference, as demonstrated in various studies \cite{cohenSubImageAnomalyDetection2021,nazareArePretrainedCNNs2018,napoletanoAnomalyDetectionNanofibrous2018, cohenSetFeaturesFinegrained2023}. One prominent example, PatchCore \cite{rothTotalRecallIndustrial2022}, utilizes a large pre-trained network to extract normal features and applies a core-set downsampling technique to mitigate storage and inference costs. This approach has exhibited strong performance and robustness across diverse industrial anomaly detection datasets. Despite these advantages, the costs associated with inference rise incrementally as the training sample size increases, due to the expanding memory bank.
\subsection{Hard Mining Strategies}
Hard mining strategies are widely employed in reconstruction methods to enhance model performance. The core idea behind these strategies is to focus the model's optimization efforts on the regions of the image that are difficult to reconstructt, typically the foreground regions. IKD \cite{caoInformativeKnowledgeDistillation2022a} introduces a hard mining method that adaptively adjusts the margin based on the statistical information (mean and variance) of feature distances within each mini-batch. This method uses a fixed discard rate as a hyperparameter, discarding feature points with smaller reconstruction errors. ReContrast \cite{guoReContrastDomainspecificAnomaly2023} refines this approach by making the discard rate variable, decreasing it over the course of iterations to achieve a coarse-to-fine optimization process. Instead of discarding the feature points themselves, ReContras discards the gradients associated with those points. EfficientAD \cite{xiaoSurveyLabelEfficientDeep2023a} takes a more aggressive approach by discarding 99.9\% of the feature points in the feature map, focusing exclusively on fine-grained optimization.

\begin{figure}[t]
  \centering
  \includegraphics[width=0.45\textwidth]{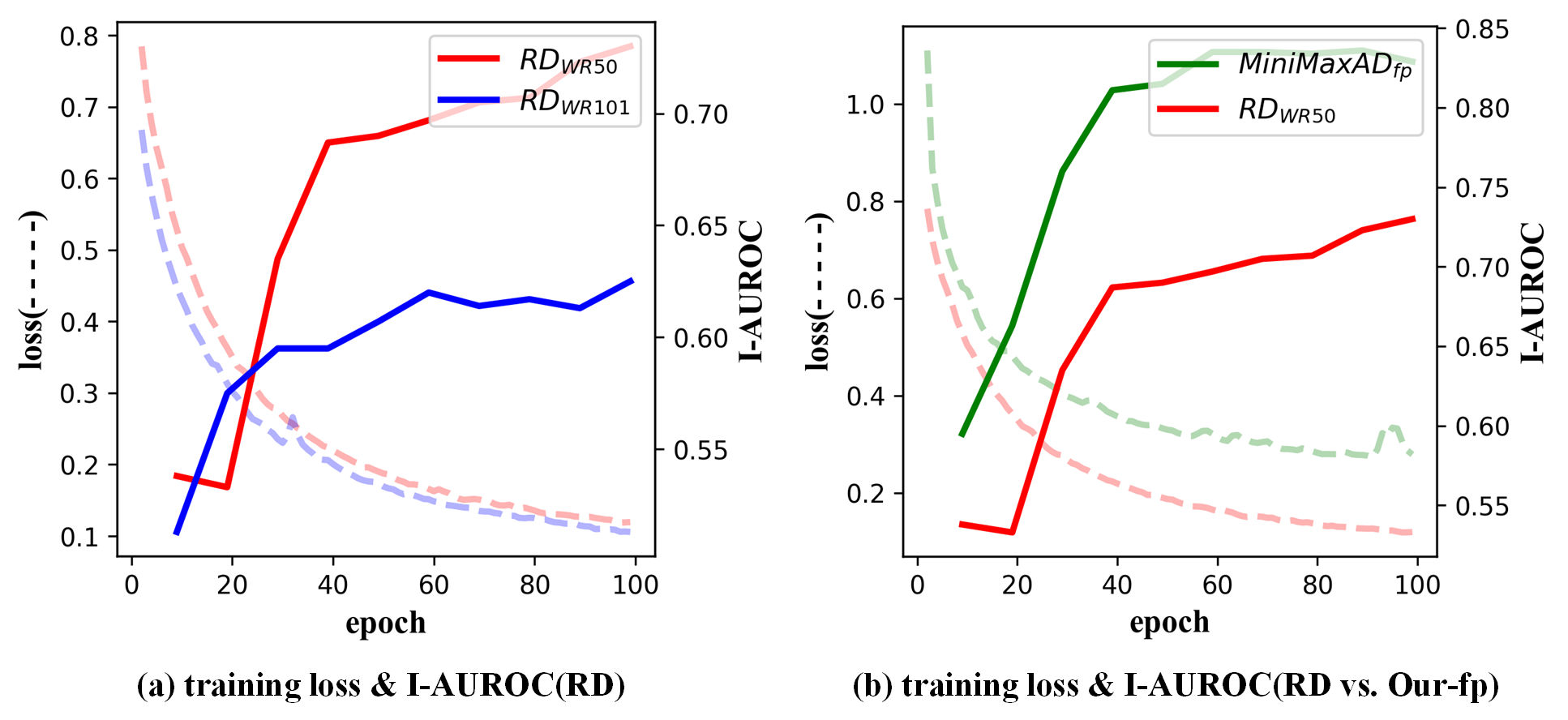}
  \caption{\textbf{(a)} Performance comparison of RD with different backbones for the \textit{drink can} scenario in the GoodsAD dataset. \textbf{(b)} Performance comparison between RD and MiniMaxAD for the \textit{drink can} scenario. }
  \label{fig3}
\end{figure}

\section{Method}
\label{sec:method}
Our study is based on the Reverse Distillation (RD) framework, which is broadly a multi-scale autoencoder. We will first summarize the classic RD methods, and use our core motivation to guide our detailed explanation of the complete MiniMaxAD methodology.

\subsection{Reverse Distillation for Anomaly Detection}
Reverse Distillation (RD) \cite{dengAnomalyDetectionReverse2022} is an important method in UAD, featuring a streamlined architecture that consists of a frozen pre-training encoder (teacher), a bottleneck, and a feature reconstruction decoder (student). The encoder extracts multi-scale spatial features from the input image, which the bottleneck then condenses to minimize redundancy. The decoder, designed as an inverse counterpart to the encoder, reconstructs the encoder’s output to achieve feature-level alignment. When faced with abnormal samples during testing, the decoder's inability to accurately reconstruct the encoder's output serves to highlight discrepancies, thereby generating an anomaly map. Formally, let \(f_E^k, f_D^k \in \mathbb{R}^{C_k \times H_k \times W_k}\) represent the feature maps output by the encoder and decoder at the \( k \)-th level, respectively. Here, \( C_k \), \( H_k \), and \( W_k \) denote the number of channels, the height, and the width of the output feature map at each level. The training objective is to minimize the global cosine distance loss between the corresponding output features of each layer for normal images, quantified by the following formula:
\begin{equation}
  \label{eq2}
  \mathcal{L}_{global}=\sum_{k=1}^3 \left(1-\frac{\mathcal{F}\left(f_E^k\right)^T\cdot\mathcal{ F}\left(f_D^k\right)}{\left\|\mathcal{F}\left(f_E^k\right)\right\|\left\|\mathcal{F}\left(f_D^k\right)\right\|}\right)
\end{equation}
Where $\mathcal{F}$ represents the flattening operation. For the tensors $f_E^k$ and $f_D^k$, point-by-point differences are calculated using the local cosine distance. This process generates a two-dimensional anomaly map, denoted by $\mathcal{M}^k \in \mathbb{R}^{H_k \times W_k}$.
\begin{equation}
  \mathcal{M}^{k}\left(h,w\right)=1-\frac{f_{E}^{k}\left(h,w\right)^{T}\cdot f_{D}^{k}\left(h,w\right)}{\left\|f_{E}^{k}\left(h,w\right)\right\|\left\|f_{D}^{k}\left(h,w\right)\right\|}
\end{equation}
Ultimately, $\mathcal{M}^k$ is upsampled to match the height and width of the original image, $I \in \mathbb{R}^{C_0 \times H_0 \times W_0}$, using an operation represented by $\Psi$. The final anomaly score map, $\mathcal{S}(h, w) \in \mathbb{R}^{H_0 \times W_0}$, is then constructed by aggregating pixel values across three layers of sub-anomaly maps:
\begin{equation}
  \mathcal{S}\left( {h,w} \right) = \sum\limits_{k = 0}^3 {\Psi \left( {{{\cal M}^k}} \right)}
\end{equation}

\begin{figure*}[t]
  \centerline{\includegraphics[width=0.9\textwidth]{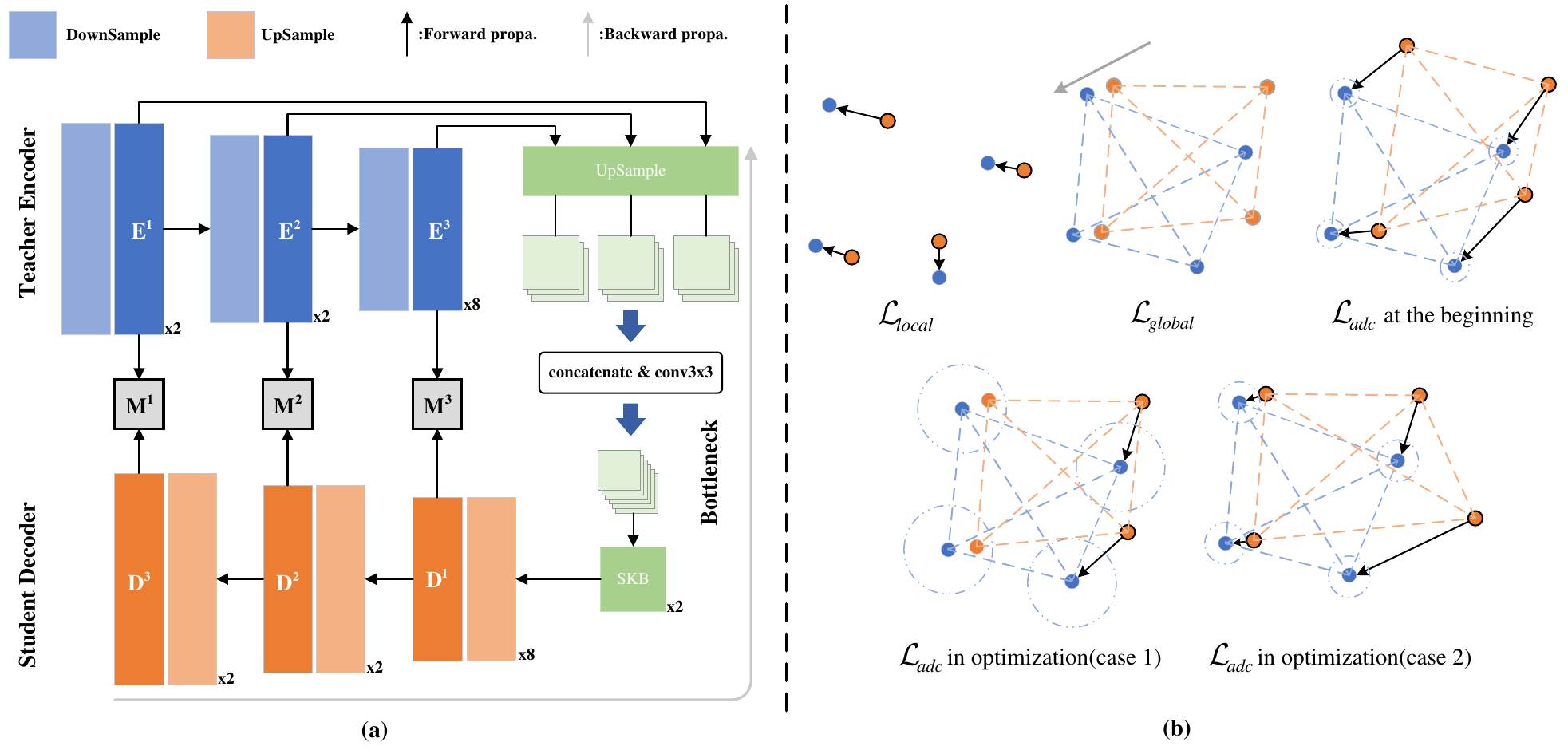}}
  \caption{\textbf{(a) Overview of Proposed MiniMaxAD.} \textbf{(b) Optimization Process of the Loss Function:} The diagram illustrates the optimization direction using arrows. Within the blue dashed circle, the yellow circle will not be optimized, i.e., the gradient will be stopped. As overall alignment improves, the radius of the blue dashed circle increases. During the optimization of \(\mathcal{L}_{adc}\), moving a single yellow circle outward in the lower right corner of Case 1 causes the blue dashed circle to contract, leading to Case 2. This adjustment encourages additional yellow circles to enter the optimization process.}
  \label{fig2}
\end{figure*}

\subsection{Expanding the Effective Capacity Cap}
The Global Response Normalization (GRN) \cite{wooConvNeXtV2CoDesigning2023} can reactivate saturated or dead neurons, thereby increasing neuron utilization and enhancing model capacity. Specifically, the GRN unit boosts channel contrast and selectivity by aggregating features at the channel level. This is achieved using the L2-norm on the feature map $X_i$, followed by normalization to compute the feature normalization coefficient $n_x$:
\begin{equation}
  n_x = \frac{g_x}{\mathbf{E}[g_x]}\in\mathcal{R^C}
\end{equation}
where $g_x = \{\|X_1\|,\|X_2\|,\ldots,\|X_C\|\}\in{\mathcal R}^C$, and $\mathbf{E}[\cdot]$ represents the computation of the mean. This equation determines the relative importance of each channel, fostering a competitive relationship among them through mutual suppression. The computed feature normalization coefficients $n_x$ are then used to calibrate the raw input:
\begin{equation}
  X = \gamma *(X * n_x) + \beta + X
\end{equation}
where \(\gamma\) and \(\beta\) are learnable hyperparameters.
\subsection{MinMax Thinking}

Our core motivation can be addressed in two ways: (1) to mitigate the ``identical shortcut'' phenomenon by reducing the number of parameters, and (2) to maximize the effective capacity of a lightweight autoencoder under the constraints imposed by (1). Here, ``effective'' highlights the model's ability to accurately reconstruct normal foreground regions, while ``capacity'' refers to the model's enhanced information density. Based on this premise, an intuitive strategy is to extract deeper, more abstract representations to achieve highly compact information storage. Consequently, large kernel convolution emerges as a powerful method for implementing this strategy. In recent visual recognition research, advances in large kernel convolution have renewed interest in convolutional networks. UniRepLKNet \cite{dingUniRepLKNetUniversalPerception2023} proposes a series of design guidelines for large-kernel convolutions; they have developed a powerful backbone equipped with a GRN Unit that perfectly fits our methodology. This network comprises several downsampling blocks and two distinct types of blocks---the Large Kernel Block (LarK Block) and the Small Kernel Block (SmaK Block). As shown in the encoder section of the Figure~\ref{fig2}(a). Specifically, \(E^1\) uses a 3$\times$3 convolution kernel in the SmaK Block, while \(E^2\) and \(E^3\) utilize a 13$\times$13 convolution kernel in the LarK Block. The full architecture of the proposed model is depicted in Figure~\ref{fig2}(a).

\begin{figure}[t]
  \centerline{\includegraphics[width=0.45\textwidth]{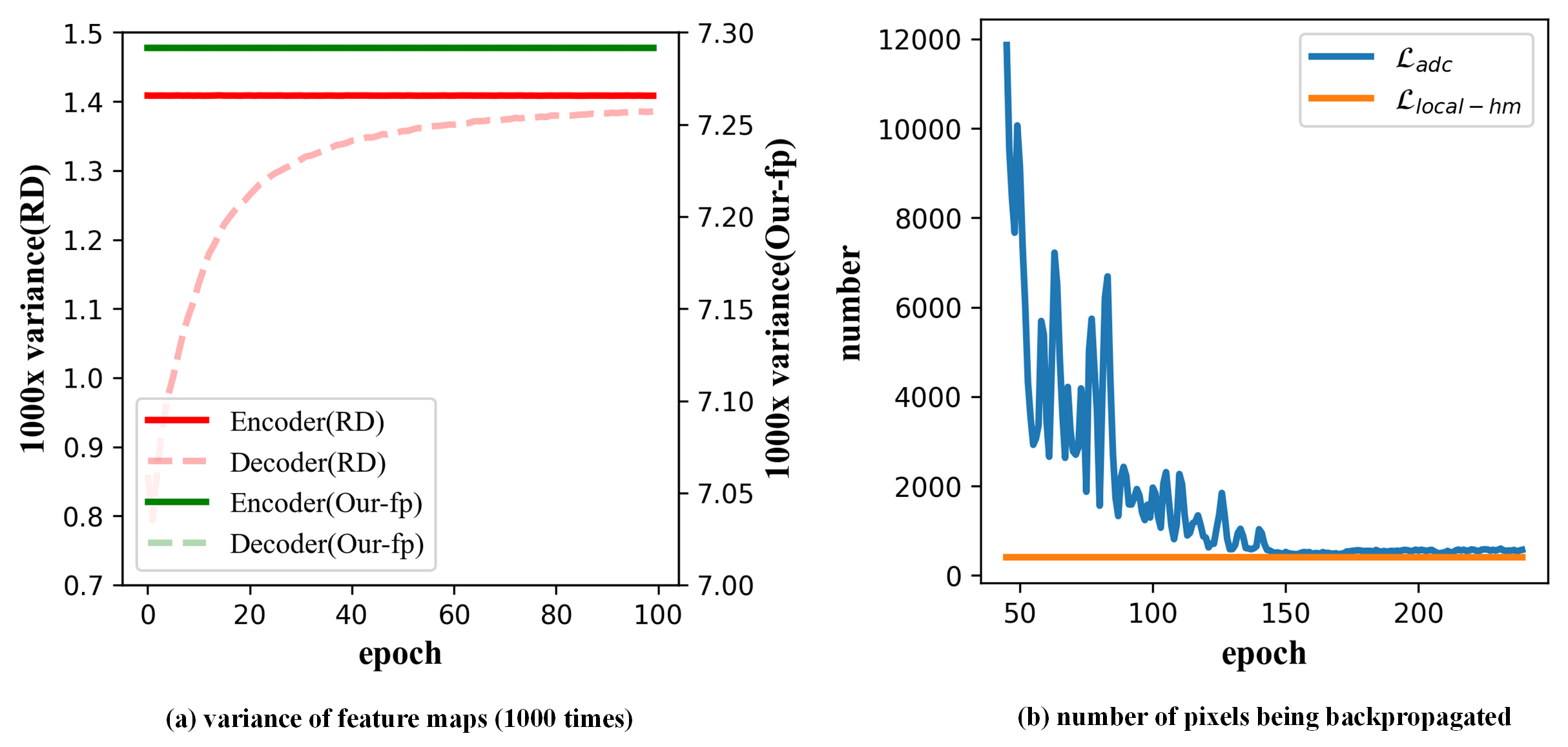}}
  \caption{\textbf{(a)} Performance comparison of RD with different backbones for the \textit{drink can} scenario in the GoodsAD dataset. \textbf{(b)} Performance comparison between RD and MiniMaxAD for the \textit{drink can} scenario. }
  \label{fig4}
\end{figure}

To illustrate the substantial advantages of MiniMaxAD over RD in handling FRADs, we conducted a quantitative analysis of the feature map alignment process, where the Decoder reconstructs the Encoder's features. We perform the L2-norm on the feature maps of the encoder and decoder, after which the variance was calculated as $\frac{1}{3}\sum_{k=1}^3\sigma({f^k_E/\left\|f^k_E\right\|})^2$ and $\frac{1}{3}\sum_{k=1}^3\sigma({f^k_D/\left\|f^k_D\right\|})^2$. As depicted in Figure~\ref{fig4}(a), benefiting from the GRN unit and large kernel design, the variance of MiniMaxAD is about five times that of RD, indicating that MiniMaxAD can characterize richer information. We conduct a more analysis of this process from the perspective of entropy, detailed in the Appendix. Additionally, Figure~\ref{fig3}(b) shows that while RD's global cosine distance loss is lower than MiniMaxAD's, RD's Image-level AUROC is significantly worse. This suggests that the reduced feature diversity in WideResNet-50 simplifies the decoder's alignment task, whereas MiniMaxAD's greater feature diversity better supports complex anomaly detection.

\subsection{Adaptive Contraction Strategy}
To be more robust, we employ a fully adaptive discard rate. First, we directly optimize the final anomaly map based on the point-by-point distance, $\mathcal{S}(h, w) \in \mathbb{R}^{H_0 \times W_0}$:
\begin{equation}
  {{\mathcal{L}}_{local}}=\frac{1}{HW}\sum\limits_{h=1}^{H}{\sum\limits_{w=1}^{W} \mathcal{S}\left( h,w \right) ^{2}}
\end{equation}

Our newly introduced \(\mathcal{L}_{local}\) is conceptually similar to applying an MSE loss between the zero tensor and the final anomaly map, \(\mathcal{S}(h, w)\). Our approach is founded on the key observation that the value of the anomaly map, $\mathcal{S}$, decreases as the alignment of the feature map improves. Furthermore, this decrease should occur in the opposite direction to the increase in the discard rate, with equivalent magnitude. We have leveraged this relationship to extract numerical features from $\mathcal{S}$. Specifically, We define \(p_{hard}\) as the primary mining factor and \(p_{lim}\) as the lower bound mining factor. Consider \(\alpha\) to be the \(p_{hard}\)-quantile of the elements in \(\mathcal{S}(h, w)\), and \(\beta\) to be the \(p_{lim}\)-quantile of the squared elements, \(\mathcal{S}(h, w)^2\). Additionally, let \(\sigma\) represent the standard deviation of \(\mathcal{S}(h, w)\). The number of pixels satisfying \(\mathcal{S}(h, w)^2 \geq \alpha - \sigma^2\) is denoted by \(\mathcal{A}\). The number of pixels influenced by the lower bound mining factor \(p_{lim}\) is indicated by \(\mathcal{B}\), calculated as \(B \times H_0 \times W_0 \times (1-p_{lim})\), where \(B\) is the batch size. We also define a mean operation, \(\mathbf{E}(\mathcal{S}|\mathcal{C})\), where elements that do not satisfy the condition \(\mathcal{C}\) will stop the gradient and not participate in the mean computation. Our proposed ADCLoss is structured as follows:
\begin{equation}
  {\mathcal{L}}_{adc}=
  \begin{cases}
    \mathbf{E}{{ \left( \mathcal{S}^{2} \mid {\mathcal{S}^2} \geq \alpha - \sigma^2 \right)}}, & \text{if $\mathcal{A} \geq \mathcal{B}$} \\

    \mathbf{E}{{ \left( \mathcal{S}^{2} \mid {\mathcal{S}^2} \geq \beta - \sigma^2 \right)}},  & \text{if $\mathcal{A} < \mathcal{B}$}
  \end{cases}
\end{equation}

This adaptive contraction strategy ensures that the property of global sensing is maintained throughout the entire optimization process of \(\mathcal{L}_{adc}\). This adaptive contraction process is illustrated in Figure~\ref{fig4}(a). By default, \(p_{hard}\) is set to 0.9999 and \(p_{lim}\) to 0.9995. The optimization process of \(\mathcal{L}_{adc}\) is depicted in Figure~\ref{fig2}(b). We characterize the loss function of a fixed mining strategy as:
\begin{equation}
  {\mathcal{L}}_{local-hm}=
  \mathbf{E}{{ \left( \mathcal{S}^{2} \mid {\mathcal{S}^2} \geq \beta \right)}}
\end{equation}

Ultimately, we provide two variants of our model: MiniMaxAD-fr and MiniMaxAD-fp. MiniMaxAD-fr is equipped with \(\mathcal{L}_{adc}\) specifically for processing FRADs, while MiniMaxAD-fp uses \(\mathcal{L}_{global}\) to handle FPADs.

\begin{table}[t]
    \centering
      \begin{tabular}{l|ccccc}
        \toprule
                & PCore & RD   & MMR           & Ours-fp/Ours-fr$^\ast$ \\
        \midrule
        I-AUROC & 71.0  & 81.0 & 84.7          & \textbf{87.4}/87.2     \\
        AUPRO   & 87.8  & 85.6 & \textbf{89.1} & 79.9/87.4              \\
        \bottomrule
      \end{tabular}
    \caption{Quantitative Results on AeBAD-S for single-class setting.}
    \label{tab1}
  \end{table}
\begin{table}[t]
  \centering
      \begin{tabular}{l|cccccc}
        \toprule
                & PCore & RD   & EAD           & MambaAD & Ours-fp \\
        \midrule
        I-AUROC & 95.1  & 96.0 & \textbf{98.1} & 94.7    & 96.1    \\
        AUPRO   & 91.2  & 70.9 & \textbf{94.0} & 91.6    & 90.4    \\
        \bottomrule
      \end{tabular}
  \caption{Quantitative Results on VisA for single-class setting.}
  \label{tab2}
\end{table}

\section{Experiments}
\label{sec:exp}

\begin{table*}[t]
  \centering
  \caption{Quantitative results on different feature-rich anomaly detection datasets.}
  \resizebox{\linewidth}{!}{
    \begin{tabular}{cccccccccc}
      \toprule
      \multirow{2}[2]{*}{Dateset} & \multirow{2}[2]{*}{Method} & \multicolumn{3}{c}{Image-level} & \multicolumn{4}{c}{Pixel-level} & \multirow{2}[2]{*}{\textbf{mAD}}                                                                                                \\
      \cmidrule(r){3-5} \cmidrule(l){6-9}
                                  &                            & AUROC                           & AP                              & F1\_max                          & AUROC            & AP               & F1\_max          & AUPRO                               \\
      \hline
      \multirow{7}[0]{*}{\parbox{2cm}{MVTec-AD                                                                                                                                                                                                                       \\ \small (multi-class)}} & RD & 94.6  & 96.5  & 95.2  & 96.1  & 48.6  & 53.8  & {91.1} &82.3\\
                                  & UniAD                      & 96.5                            & 98.8                            & 96.2                             & 96.8             & 43.4             & 49.5             & 90.7             & 81.7             \\
                                  & SimpleNet                  & 95.3                            & 98.4                            & 95.8                             & \underline{96.9} & 45.9             & 49.7             & 86.5             & 81.2             \\
                                  & DiAD                       & 97.2                            & \underline{99.0}                & 96.5                             & 96.8             & 52.6             & {55.5}           & 90.7             & {84.0}           \\
                                  & MambaAD                    & \underline{98.6}                & \textbf{99.6}                   & \textbf{97.8}                    & \textbf{97.7}    & \underline{56.3} & \underline{59.2} & \textbf{93.1}    & \underline{86.0} \\
                                  & MiniMaxAD-fr(Ours)         & \textbf{98.8}                   & \textbf{99.6}                   & \underline{97.7}                 & 96.1             & \textbf{58.5}    & \textbf{59.5}    & \underline{92.2} & \textbf{86.1}    \\
      \hline
      \multirow{7}[0]{*}{\parbox{2cm}{VisA                                                                                                                                                                                                                           \\ \small (multi-class)}} & RD & {92.4}  & {92.4}  & \underline{89.6} & 98.1 & 38.0  & 42.6  & \textbf{91.8}& 77.8 \\
                                  & UniAD                      & 88.8                            & 90.8                            & 85.8                             & \underline{98.3} & 33.7             & 39.0             & 85.5             & 74.6             \\
                                  & SimpleNet                  & 87.2                            & 87.0                            & 81.8                             & 96.8             & 34.7             & 37.8             & 81.4             & 72.4             \\
                                  & DiAD                       & 86.8                            & 88.3                            & 85.1                             & 96.0             & 26.1             & 33.0             & 75.2             & 70.1             \\
                                  & MambaAD                    & \underline{94.3}                & \underline{94.5}                & {89.4}                           & \textbf{98.5}    & 39.4             & \underline{44.0} & \underline{91.0} & \underline{78.7} \\
                                  & MiniMaxAD-fr(Ours)         & \textbf{96.9}                   & \textbf{97.2}                   & \textbf{94.1}                    & 97.4             & \textbf{48.6}    & \textbf{51.5}    & {90.1}           & \textbf{82.3}    \\
      \hline
      \multirow{7}[0]{*}{\parbox{2cm}{Real-IAD                                                                                                                                                                                                                       \\ \small (multi-class)}}& RD&82.4  & 79.0  & 73.9  & 97.3  & 25.0  & 32.7  & \underline{89.6} &\underline{68.6} \\
                                  & UniAD                      & 83.0                            & 80.9                            & 74.3                             & 97.3             & 21.1             & 29.2             & 86.7             & 67.5             \\
                                  & SimpleNet                  & 57.2                            & 53.4                            & 61.5                             & 75.7             & \;\;2.8          & \;\;6.5          & 39.0             & 42.3             \\
                                  & DiAD                       & 75.6                            & 66.4                            & 69.9                             & 88.0             & \;\;2.9          & \;\;7.1          & 58.1             & 52.6             \\
                                  & MambaAD                    & \textbf{86.3}                   & \textbf{84.6}                   & \textbf{77.0}                    & \textbf{98.5}    & \underline{33.0} & \underline{38.7} & \textbf{90.5}    & \textbf{72.7}    \\
                                  & MiniMaxAD-fr(Ours)         & 85.1                            & \underline{84.1}                & \underline{75.8}                 & 96.9             & \textbf{36.0}    & \textbf{42.1}    & 88.8             & \textbf{72.7}    \\
      \hline
      \multirow{6}[0]{*}{\parbox{2cm}{Real-IAD                                                                                                                                                                                                                       \\ \small (single-class)}}
                                  & PatchCore                  & 90.3                            & \underline{88.6}                & \underline{81.3}                 & 98.2             & 36.8             & 41.3             & 89.8             & 75.2             \\
                                  & RD                         & 89.5                            & 87.9                            & 80.5                             & \underline{98.6} & \underline{41.0} & \underline{46.0} & \textbf{93.3}    & \underline{76.7} \\
                                  & UniAD                      & 81.6                            & 77.3                            & 73.4                             & 97.6             & 17.9             & 25.1             & 86.9             & 65.7             \\
                                  & SimpleNet                  & 88.9                            & 87.4                            & 80.4                             & 96.8             & 20.8             & 28.8             & 83.3             & 69.5             \\
                                  & MVAD                       & \underline{90.2}                & 88.2                            & 81.0                             & \textbf{98.9}    & 34.6             & 39.9             & \underline{92.3} & 75.0             \\
                                  & MiniMaxAD-fr(Ours)         & \textbf{91.2}                   & \textbf{89.7}                   & \textbf{82.4}                    & 98.3             & \textbf{44.2}    & \textbf{47.6}    & \underline{92.3} & \textbf{80.0}    \\
      \hline
      \multirow{5}[0]{*}{\parbox{2cm}{GoodsAD                                                                                                                                                                                                                        \\ \small (single-class)}}
                                  & PatchCore                  & \underline{85.2}                & \underline{85.9}                & \textbf{82.9}                    & \textbf{97.2}    & \textbf{49.4}    & \textbf{53.0}    & \underline{84.9} & \textbf{76.9}    \\
                                  & RD                         & 67.7                            & 68.9                            & 74.1                             & 93.3             & 16.4             & 23.8             & 79.0             & 60.5             \\
                                  & SimpleNet                  & 74.1                            & 76.9                            & 75.5                             & 86.9             & 21.9             & 27.8             & 65.2             & 61.2             \\
                                  & MiniMaxAD-fr(Ours)         & \textbf{86.2}                   & \textbf{86.9}                   & \underline{82.7}                 & \underline{96.2} & \underline{43.9} & \underline{46.9} & \textbf{86.0}    & \underline{75.5} \\
      \bottomrule
    \end{tabular}
  }
  \label{tab:results_pr}
\end{table*}

\begin{figure*}[t]
  \centerline{\includegraphics[width=0.7\linewidth]{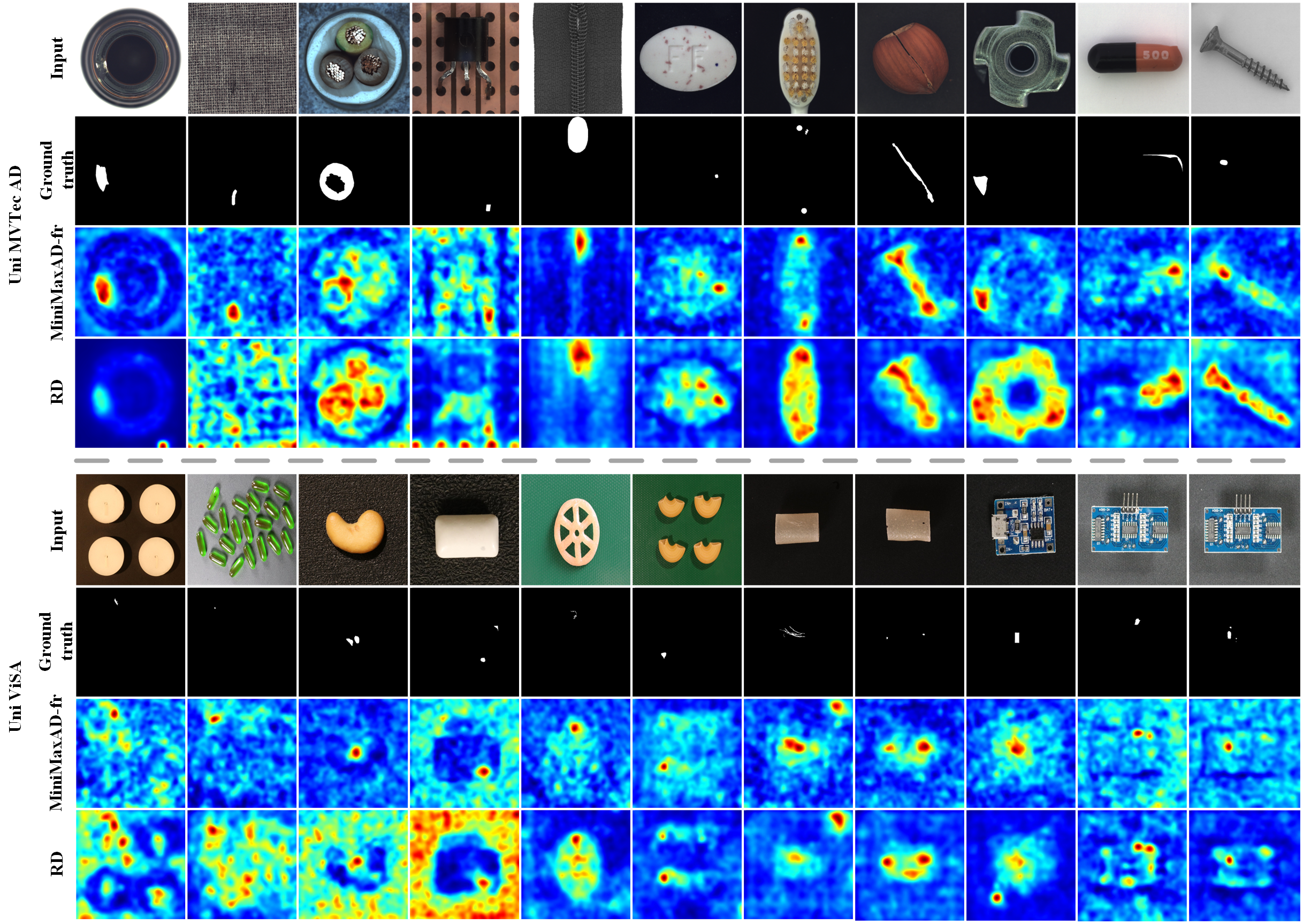}}
  \caption{Visualization of qualitative results for the MVTecAD dataset in the multi-class setting.}
  \label{figa2}
\end{figure*}

\subsection{Experimental Settings}
\label{Settings}
\textbf{Datasets.}
\textbf{MVTec AD} and \textbf{VisA} feature high-resolution images across multiple object and texture categories, highlighting both surface and structural defects. They are the two most popular datasets for anomaly detection, covering a wide range of anomalies at different scales in scenarios such as components, food, and medical care. \textbf{AeBAD-S} is specifically focused on domain diversity within aero-engine blade images.The \textbf{GoodsAD} dataset addresses product damage detection in unmanned supermarket scenarios. It captures extensive intra-class diversity with 6,124 high-resolution images of common supermarket goods, showcasing a single product offered by multiple brands. \textbf{Real-IAD} is a recent, large {industrial anomaly detection} dataset comprising 150K images across 30 categories, with each category containing five default perspectives. Detailed information is presented in Appendix.

\textbf{Implementation.} Network optimization was carried out using an AdamW optimizer \cite{loshchilovDecoupledWeightDecay2019}, with the $\beta$ parameters set to (0.9, 0.999) and a weight decay of 0.00005. The learning rate was set at 0.005, accompanied by an ExponentialLR scheduler with a gamma of 0.995. The batch size for all experiments was 16. For experiments in the single-class setting, GoodsAD was trained for 240 epochs, ViSA for 150 epochs, AeBAD-S for 130 epochs, and Real-IAD for 50 epochs. In the multi-class setting, MVTec AD and ViSA were trained for 160 epochs, while Real-IAD was trained for only 1 epoch. Results were reported from the \textbf{last epoch} for each scenario. All experiments were performed on an NVIDIA RTX 4090 GPU using PyTorch 2.0. The model does not use efficient large kernel convolution as suggested by \cite{dingUniRepLKNetUniversalPerception2023}.

\textbf{Metrics.} We use two of the most commonly employed metrics for anomaly detection and segmentation as our basic evaluation metrics \cite{bergmannMVTecADComprehensive2019}: Area Under the Receiver Operating Characteristic Curve (AUROC) and pixel-level Area Under the Per-Region Overlap (AUPRO). Additionally, following previous work \cite{zhangExploringPlainViT2023a,heMambaADExploringState2024}, we adopt supplementary metrics on certain datasets for a more comprehensive evaluation, including Average Precision (AP) \cite{zavrtanikDRAEMDiscriminativelyTrained2021} and F1-score-max (F1\_max) \cite{zouSPotthedifferenceSelfsupervisedPretraining2022}. We also calculate the average of the seven evaluation metrics mentioned above, denoted as mAD, to represent the model's overall performance.

\subsection{Anomaly Detection and Segmentation on FPADs}

\begin{table}[ht]
  \centering
    \captionof{table}{Ablation study on structural elements of the model.}
    \begin{tabular}{cccc}
      \toprule
      Encoder    & Decoder    & GRN       & I-AUROC          \\
      \midrule
      LarK Block & LarK Block & \ding{52} & \textbf{95.5}    \\
      LarK Block & LarK Block & \ding{56} & 94.5             \\
      LarK Block & SarK Block & \ding{52} & \underline{94.7} \\
      LarK Block & SarK Block & \ding{56} & 93.7             \\
      SarK Block & LarK Block & \ding{52} & 93.5             \\
      SarK Block & LarK Block & \ding{56} & 93.8             \\
      SarK Block & SarK Block & \ding{52} & 93.7             \\
      SarK Block & SarK Block & \ding{56} & 93.5             \\
      \bottomrule
    \end{tabular}
    \label{tab:ab2}
\end{table}

In this section, we focus on demonstrating the performance of MiniMaxAD on FPADs using the two most commonly used metrics (see Section~\ref{Settings}). We compare our method with RD \cite{dengAnomalyDetectionReverse2022}, PatchCore \cite{ishidaSAPatchCoreAnomalyDetection2023}, MMR \cite{zhangIndustrialAnomalyDetection2023}, EfficientAD \cite{batznerEfficientADAccurateVisual2023}, and MambaAD \cite{heMambaADExploringState2024}, all of which are powerful state-of-the-art (SoTA) methods.

As shown in the table~\ref{tab1}, on the AeBAD-S, our model (Ours-fp) exhibits a 2.7$\uparrow$ increase in classification and a 9.2$\downarrow$ decrease in segmentation compared to MMR. We hypothesize that this sharp decline in segmentation performance is related to the domain shift in the AeBAD-S. However, after adding all 15 classes of training samples from MVTecAD into its training set to convert it into FRAD and training MiniMax-fr for 20 epochs (denoted as Ours-fr$^\ast$). This adjustment significantly improved segmentation performance, albeit with a slight decrease in detection performance. Specifically, compared to MMR, the performance changed by 2.5$\uparrow$ in classification and 1.7$\downarrow$ in segmentation. We view the addition of irrelevant samples to the training set as a form of \textbf{strong regularization}, which helps prevent model overfitting by reducing excessive focus on specific regions or a priori domains. In addition, Table ~\ref{tab2} demonstrates the gap between MiniMax-fp and SoTA methods, with our method achieving a significant improvement of 19.5$\uparrow$ in segmentation performance over the baseline method RD.

It is important to emphasize that anomaly detection for FPADs is not the primary focus of our research. While some methods, such as MambaAD, perform exceptionally well on FRADs, they may still struggle with FPADs. This limitation precisely motivates our proposed classification approach, which aims to identify targeted solutions based on different application scenarios.

\begin{figure*}[t]
  \centerline{\includegraphics[width=0.7\linewidth]{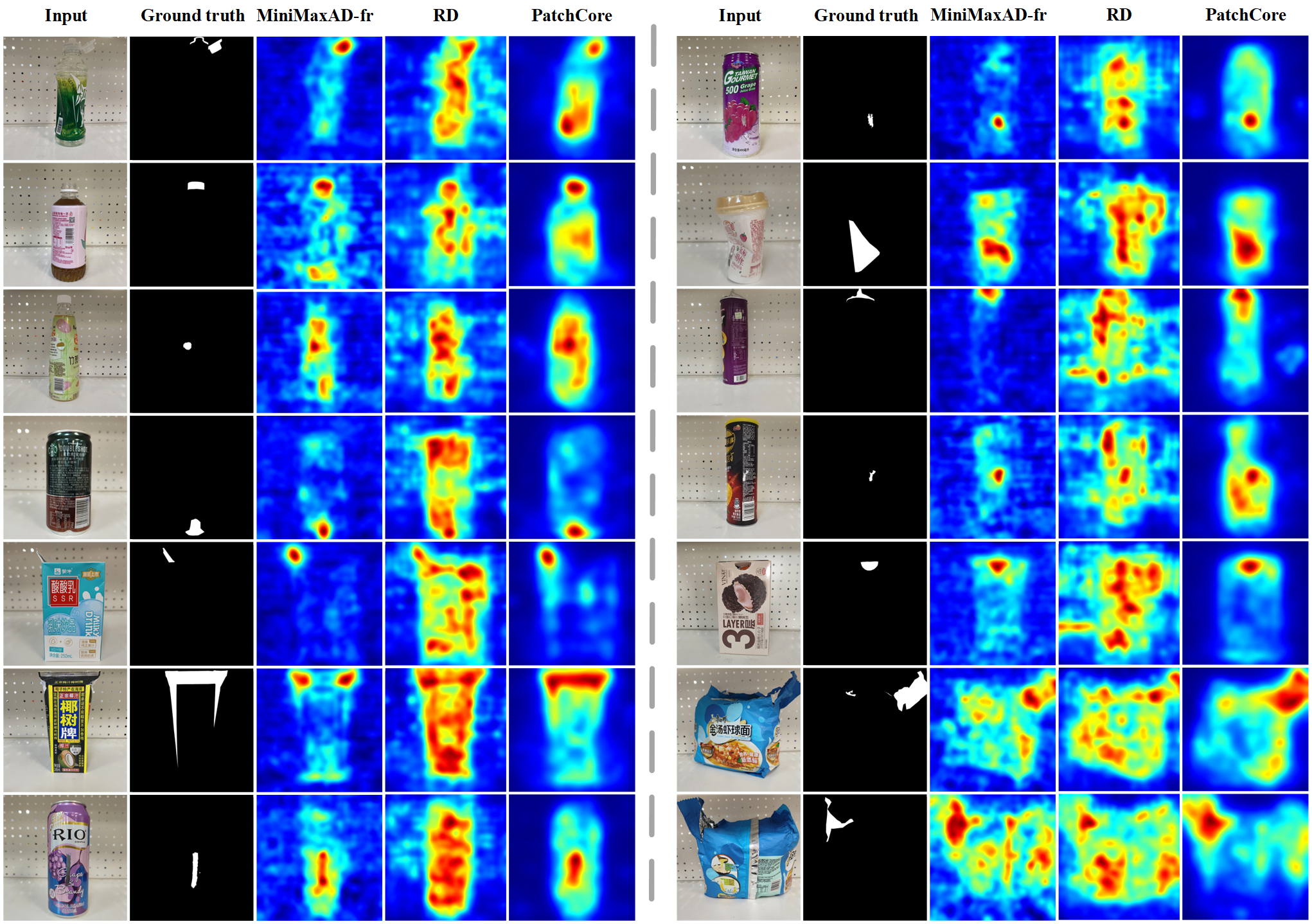}}
  \caption{Visualization of qualitative results for the GoodsAD dataset in the single-class setting.}
  \label{figa1}
\end{figure*}

\begin{table*}[!t]
  \centering
  \caption{ Ablation study on loss functions and dataset classification methods, measured in I-AUROC(\%). }
  \begin{tabular}{l|l|ccc|cc}
    \toprule
             &                                 & \multicolumn{3}{c|}{FRADs} & \multicolumn{2}{c}{FPADs}                                                                \\
             &                                 & GoodsAD                    & Uni ViSA                  & Uni AeBAD-S\&MVTecAD         & AeBAD-S       & ViSA          \\
    \midrule
    Baseline & RD + $\mathcal{L}_{global}$     & 66.5                       & 92.7                      & 76.2\&89.6                   & 81.0          & 96.0          \\
             & Ours + $\mathcal{L}_{local}$    & 80.1                       & 93.0                      & 84.3\&97.1                   & 85.1          & 93.8          \\
             & Ours + $\mathcal{L}_{local-hm}$ & 80.6                       & 96.3                      & 86.2\&98.2                   & 85.4          & 96.0          \\
    Ours-fp  & Ours + $\mathcal{L}_{global}$   & 77.5                       & 93.1                      & 84.9\&96.3                   & \textbf{87.4} & \textbf{96.1} \\
    Ours-fr  & Ours + $\mathcal{L}_{adc}$      & \textbf{86.1}              & \textbf{96.9}             & \textbf{87.2}\&\textbf{98.4} & 83.9          & 95.6          \\
    \bottomrule
  \end{tabular}
  \label{tab:ab1}
\end{table*}

\subsection{Anomaly Detection and Segmentation on FRADs}

\begin{table*}[t]
  \centering
  \caption{{Ablation study on on the pre-trained backbone. For the depth column, ``9+9'' means 9 LarK Blocks and 9 SmaK Blocks. Our model uses UniRepLKNet-N by default.}}
  \resizebox{\linewidth}{!}{
    \begin{tabular}{clccccccccc}
      \toprule
      \multirow{2}[2]{*}{Backbone} & \multirow{2}[2]{*}{Depth} & \multirow{2}[2]{*}{Params(M)} & \multicolumn{3}{c}{Image-level} & \multicolumn{4}{c}{Pixel-level} & \multirow{2}[2]{*}{\textbf{mAD}}                                         \\
      \cmidrule(r){4-6} \cmidrule(l){7-10}
                                   &                           &                               & AUROC                           & AP                              & F1\_max                          & AUROC & AP   & F1\_max & AUPRO        \\
      \midrule
      UniRepLKNet-F                & $\left[2,2,6+0 \right]$   & 10.8                          & 97.3                            & 98.7                            & 96.7                             & 94.7  & 51.3 & 54.0    & 89.5  & 83.2 \\
      UniRepLKNet-P                & $\left[2,2,6+0 \right]$   & 19.0                          & 98.1                            & 99.3                            & 97.5                             & 95.0  & 53.4 & 56.0    & 90.7  & 84.3 \\
      UniRepLKNet-N                & $\left[2,2,8+0 \right]$   & 33.1                          & 98.8                            & 99.6                            & 97.7                             & 96.1  & 58.5 & 59.5    & 92.2  & 86.1 \\
      UniRepLKNet-T                & $\left[2,2,9+9 \right]$   & 51.4                          & 99.0                            & 99.6                            & 98.0                             & 97.1  & 61.5 & 61.9    & 93.6  & 87.2 \\
      UniRepLKNet-S                & $\left[2,2,9+18\right]$   & 96.4                          & 99.1                            & 99.7                            & 98.3                             & 97.1  & 62.8 & 62.7    & 93.8  & 87.6 \\
      \bottomrule
    \end{tabular}
  }
  \label{tab:backbone}
\end{table*}

\begin{table*}[h]
  \centering
  \caption{Complexity comparison of SoTA methods. {All complexity metrics} are measured on an NVIDIA RTX 4090 with a batch size of 16 {, and the input resolution is $256\times256$.} {The CPU used is an AMD EPYC 9654 96-Core Processor with 16 vCPUs allocated.} mAD represents the mean value of mAD across the benchmark for the three multi-class settings.}
  \begin{tabular}{ccccccc}
    \toprule
    Method    & Params(M)        & FLOPs(G)          & FPS             & {Memory(M)}      & {Storage(M)}   & mAD              \\
    \midrule
    RD        & 117.1            & 30.7              & \underline{665} & 2179             & 448            & 76.2             \\
    UniAD     & \textbf{24.5}    & \textbf{ \;3.6}   & 188             & 1373             & \textbf{95}    & 74.6             \\
    SimpleNet & 72.8             & 16.1              & 52              & \textbf{1035}    & 279            & 65.3             \\
    DiAD      & 1331.3           & 451.5             & -               & -                & -              & 68.9             \\
    MambaAD   & \underline{25.7} & \;\underline{8.3} & 216             & 2309             & \underline{99} & \underline{79.1} \\
    Ours      & 33.1             & \;\underline{7.0} & \textbf{906}    & \underline{1327} & 127            & \textbf{80.4}    \\
    \bottomrule
  \end{tabular}
  \label{tab:efficient}
\end{table*}

\textbf{Quantitative results. }
We compare our method with current SoTA methods using image-level and pixel-level metrics (see Section~\ref{Settings}) across a range of datasets. This section focuses on comparisons with UniAD \cite{youUnifiedModelMulticlass2022}, DiAD \cite{heDiADDiffusionbasedFramework2023}, and MambaAD \cite{heMambaADExploringState2024}, which have been recently tailored for multi-class settings. We also compare other SoTA methods, such as RD \cite{dengAnomalyDetectionReverse2022}, SimpleNet \cite{liuSimpleNetSimpleNetwork2023} and PatchCore \cite{ishidaSAPatchCoreAnomalyDetection2023}. Additionally, a comparison with a more recent MVAD \cite{heLearningMultiviewAnomaly2024} method designed for multi-view anomaly detection is included.

As shown in Table~\ref{tab:results_pr}, our MiniMaxAD-fr outperforms all comparative methods on FRADs. Specifically, in the multi-class benchmark, our method's I-AUROC improves by 0.2$\uparrow$/2.6$\uparrow$ on the MVTecAD and VisA datasets, respectively, compared to MambaAD. On the mAD metric, we see an improvement of 0.1$\uparrow$/3.6$\uparrow$. When evaluated on the more challenging Real-IAD dataset, our mAD metrics are on par with MambaAD, both reaching 72.7. In the single-class Real-IAD benchmark, our method improves detection performance at the image level by 1.0$\uparrow$/1.5$\uparrow$/1.4$\uparrow$ and the mAD metrics by 5.0$\uparrow$ compared to MVAD, which is specifically designed for multi-view settings. Additionally, the single-class GoodsAD further demonstrates the competitiveness of our method, achieving a 1.0$\uparrow$/1.1$\uparrow$ improvement in the two fundamental metrics, I-AUROC and AUPRO, compared to PatchCore. While the mAD of MiniMaxAD-fr in GoodsAD is lower than that of PatchCore, which is challenging to implement in multi-class setting due to its memory bank mechanism, our approach remains scalable. These results fully demonstrate the validity and necessity of our approach. More detailed results are presented in the Appendix.

\textbf{Qualitative results. }
We visualize the experimental results of MiniMaxAD-fr compared to the baseline model RD in the multi-class setting to further illustrate the model's lack of ``effective capacity'', as shown in Figure~\ref{figa2}. Specifically, during the training process of the reconstruction model, the objective is to iteratively train the model to reconstruct normal samples from the training set. In the multi-class setting, sharing a single model across numerous items significantly exacerbates the model's memory burden.
Although the RD model has more than three times the number of parameters as MiniMaxAD, it struggles to accurately memorize the normal regions of objects, resulting in numerous false-negative regions in the anomaly map. More concerning, its inconsistent memory even causes the anomaly maps to display unexpected outliers, as seen in the bottom regions of the first four columns of the anomaly maps for the MVTecAD dataset in Figure~\ref{figa2}.

Figure~\ref{figa1} presents the qualitative results on GoodsAD, along with a comparison to PatchCore. As shown in the figure, PatchCore benefits from its explicit memory bank memorization mechanism, which prevents misremembering and results in smoother anomaly maps overall. However, this explicit memorization becomes a significant burden as the training set increases in size. Additionally, The figure shows that PatchCore encounters some false positives when dealing with domain shift, and the performance degradation presented in Table~\ref{tab1} directly reflects this issue.

\subsection{Ablation and Analysis}

\textbf{Comparison of the effectiveness of the model's structural elements}. Table~\ref{tab:ab2} presents ablation studies on the Uni MVTecAD dataset for GRN and large kernel designs. Specifically, we pre-trained four encoders on a sampled version of ImageNet, ImageNet-100 \cite{tianContrastiveMultiviewCoding2019}, and their validation set classification scores were nearly identical. Further details on the pre-training process are provided in Appendix. For the same architecture, models using GRN typically outperform those without GRN. Additionally, performance tends to decline and lacks significant comparability when the SarK Block is used as an encoder.

\textbf{Comparison of loss functions applied to different types of datasets.} A comprehensive ablation study is presented in Table~\ref{tab:ab1}, validating the effectiveness of the loss function and dataset taxonomy. Compared to RD, MiniMaxAD+$\mathcal{L}_{global}$ demonstrates a significant improvement in performance on FRADs and FPADs, affirming the suitability of large kernel convolutional networks with GRN units for the multi-scale reconstruction framework. Our asymptotic adaptive contraction strategy, which is tailored for FRADs based on \(\mathcal{L}_{local}\), achieves optimal performance on FRADs. In contrast, the performance of MiniMaxAD+\(\mathcal{L}_{adc}\) is degraded compared to MiniMaxAD+\(\mathcal{L}_{global}\) on FPADs. The reason for this degradation is that while \(\mathcal{L}_{adc}\) is based on \(\mathcal{L}_{local}\) and offers certain advantages, it compromises global consistency to some extent \cite{guoReContrastDomainspecificAnomaly2023}. Specifically, easier-to-fit feature-poor anomaly detection datasets (FPADs) do not benefit from \(\mathcal{L}_{adc}\), which relies on point-by-point distances, as these distances are more stringent compared to global distances. However, transforming FPADs into FRADs is straightforward. As reported in the third column of our table, we merged scenarios to unify AeBAD-S with MVTec AD for training. Here we resize MVTec AD to 224 $\times$ 224, other settings follow the default AeBAD-S. The results demonstrate that converting FPADs to FRADs using the unified setting in our methodology yields more gains than losses. The additional results in Table~\ref{tab1} further highlight the significant improvement in segmentation performance. Further ablation studies and experimental results are presented in the Appendix.

\begin{figure}[t]
  \centering
    \caption{{Comparison of the ERF of different networks.}} 
    \includegraphics[width=0.92\linewidth]{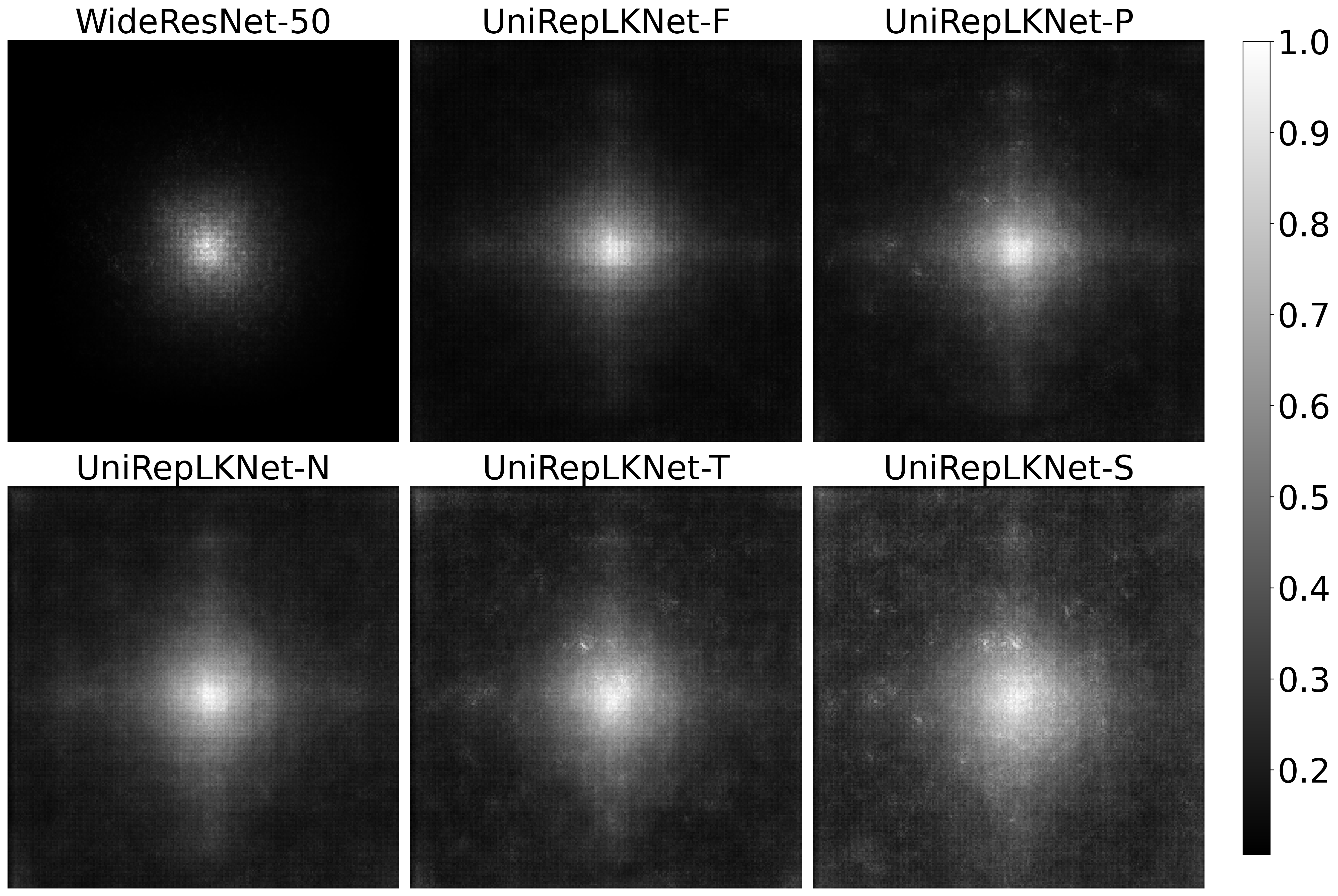} 
    \label{ERF}
\end{figure}

\textbf{Comparison of different pre-trained backbones \& receptive field analysis.} According to the Effective Receptive Field (ERF) theory \cite{luoUnderstandingEffectiveReceptive2016}, a model's effective receptive field occupies only a fraction of its theoretical receptive field and exhibits Gaussian distribution characteristics. The extent of this influence can be determined by calculating the gradient of a point on the deep feature map with respect to the original pixel. Recent studies \cite{kimDeadPixelTest2021,dingScalingYourKernels2022} have visualized ERFs by backpropagating from the center of the deep feature map. Following the visualization approach described in \cite{dingScalingYourKernels2022}, Figure~\ref{ERF} presents the ERF visualization results for various backbones \cite{dingUniRepLKNetUniversalPerception2023,zagoruykoWideResidualNetworks2017}. As shown in the figure, deeper models exhibit larger ERFs, indicating that the bottleneck layer can integrate more abstract features. Additionally, we evaluated the impact of backbones with larger ERFs on model performance using the Uni MVTecAD dataset. Across all backbones, our bottleneck layer consistently employs two Smark modules, and the decoder mirrors the encoder directly. The results in Table~\ref{tab:backbone} demonstrate that our model adheres to the ``scaling law'', suggesting that larger backbones enhance performance. Notably, even when using the larger UniRepLKNet-S as the backbone, the total number of model parameters remains lower than that of the standard RD.

\textbf{Complexity comparison of different SoTA models.} Table~\ref{tab:efficient} presents the detailed results of the complexity comparison. The architecture of MiniMaxAD inherits the high efficiency of the pure convolutional RD architecture while further reducing computational complexity. This enhancement significantly improves its throughput compared to other SoTA methods, providing a solid foundation for real-world implementation.

\section{Conclusions}
\label{sec:conclu}
This paper introduces SSFilter, which is the first method to apply sample-level filtering in noisy unsupervised anomaly detection. The advantage of this filtering approach is its high scalability. SSFilter not only achieves impressive performance in end-to-end fully unsupervised anomaly detection but also allows filtering of the entire dataset using a trained model, thus serving as a bridge between UAD and FUAD methods. We conducted extensive experiments to demonstrate the outstanding performance of our approach.
\\
\textbf{limitations.}
Despite SSFilter’s strong performance in noisy scenarios, its filtering mechanism discards some normal samples, causing a degree of performance degradation. Data-efficient unsupervised anomaly detection methods could improve our approach in future work.
{
    \small
    \bibliographystyle{ieeenat_fullname}
    \bibliography{main}
}
\clearpage
\setcounter{page}{1}
\setcounter{section}{0}
\maketitlesupplementary
\renewcommand{\thesection}{\Alph{section}}

\section{Datasets}
\label{app_datasets}

\textbf{Real-IAD} is a large-scale, real-world, multi-view dataset for industrial anomaly detection. It comprises 30 different object categories, with 99,721 normal images and 51,329 abnormal images, making it more challenging than previous datasets. all images were resized to a uniform resolution of 256 $\times$ 256.
\\
\textbf{MVTec AD} is a widely recognized industrial anomaly detection dataset, comprising over 5,000 high-resolution images across fifteen distinct object and texture categories. Each category includes a set of defect-free training images and a test set featuring a range of defects as well as defect-free images. In our experiments, all images were resized to a uniform resolution of 256 $\times$ 256.
\\
\textbf{VisA} is a challenging anomaly detection dataset with 10,821 images across 12 subsets, each representing a different class of objects. The anomalies in these images include surface defects like scratches, dents, discoloration, or cracks, and structural defects such as misaligned or missing parts. We applied the unsupervised default settings \cite{zouSPotthedifferenceSelfsupervisedPretraining2022} to segregate the training and test sets. In our experiments, all images were resized to a uniform resolution of 256 $\times$ 256.
\\
\textbf{AeBAD-S} shifts focus from the diversity of defect categories, as seen in datasets like MVTec AD and VisA, to the diversity of domains within the same category. AeBAD-S aims to enhance the automation of anomaly detection in aero-engine blades, crucial for their stable operation. This dataset features various scales of individual blade images, characterized by non-aligned samples and a domain shift between the normal sample distribution in the test set and that in the training set, primarily due to variations in illumination and viewing angles. We resized the images to 256 $\times$ 256 and then center-cropped them to 224 $\times$ 224 for our experiments.
\\
\textbf{GoodsAD} includes 6,124 images across six categories of common supermarket goods, captured at a high resolution of 3000 $\times$ 3000. Each image features a single item, typically centered, with anomalies occupying only a small fraction of the image pixels. In our experiments, all images were resized to a uniform resolution of 224 $\times$ 224.
\begin{table}[h]
  \centering
  \caption{Score of the pre-trained encoder on the validation set.}
  \begin{tabular}{cccc}
    \toprule
    Block      & GRN       & Top-1 Acc. & Top-5 Acc. \\
    \midrule
    LarK Block & \ding{52} & 83.1       & 96.0       \\
    LarK Block & \ding{56} & 82.9       & 95.9       \\
    SarK Block & \ding{52} & 82.7       & 95.9       \\
    SarK Block & \ding{56} & 82.5       & 95.9       \\
    \bottomrule
  \end{tabular}
  \label{tab:ab3}
\end{table}
\section{Additional Details}
\textbf{Data sources.} The results for the comparison methods on AeBAD-S \cite{zhangIndustrialAnomalyDetection2023} are partially from the original paper. The results of the MVTecAD, ViSA, and Real-IAD comparison tests in the multi-class setting were extracted from MambaAD \cite{heMambaADExploringState2024}. Meanwhile, the comparison data for Real-IAD in the single-class setting was partially taken from MVAD \cite{heLearningMultiviewAnomaly2024}. We are very grateful to the authors for conducting these detailed experiments.

\textbf{Pre-trained encoder.} We pre-trained four encoders on a subset of ImageNet, specifically ImageNet-100 \cite{tianContrastiveMultiviewCoding2019}. For the 100 categories in the training set, we randomly selected 100 images per category to form the validation set. Each of our models was trained for 50 epochs, and their validation set scores are presented in Table A1. At the model level, the LarK block refers to the original UniRepLKNet-N architecture, where the large kernel block is the DilatedReparamBlock with a kernel size of 13. The SarK module, on the other hand, replaces the DilatedReparamBlock with a 3x3 depth-wise convolution.

\section{Additional Experiments and Analysis}

\subsection{Alignment Process in Entropy Perspective}

\begin{figure}[t]
  \centerline{\includegraphics[width=0.4\textwidth]{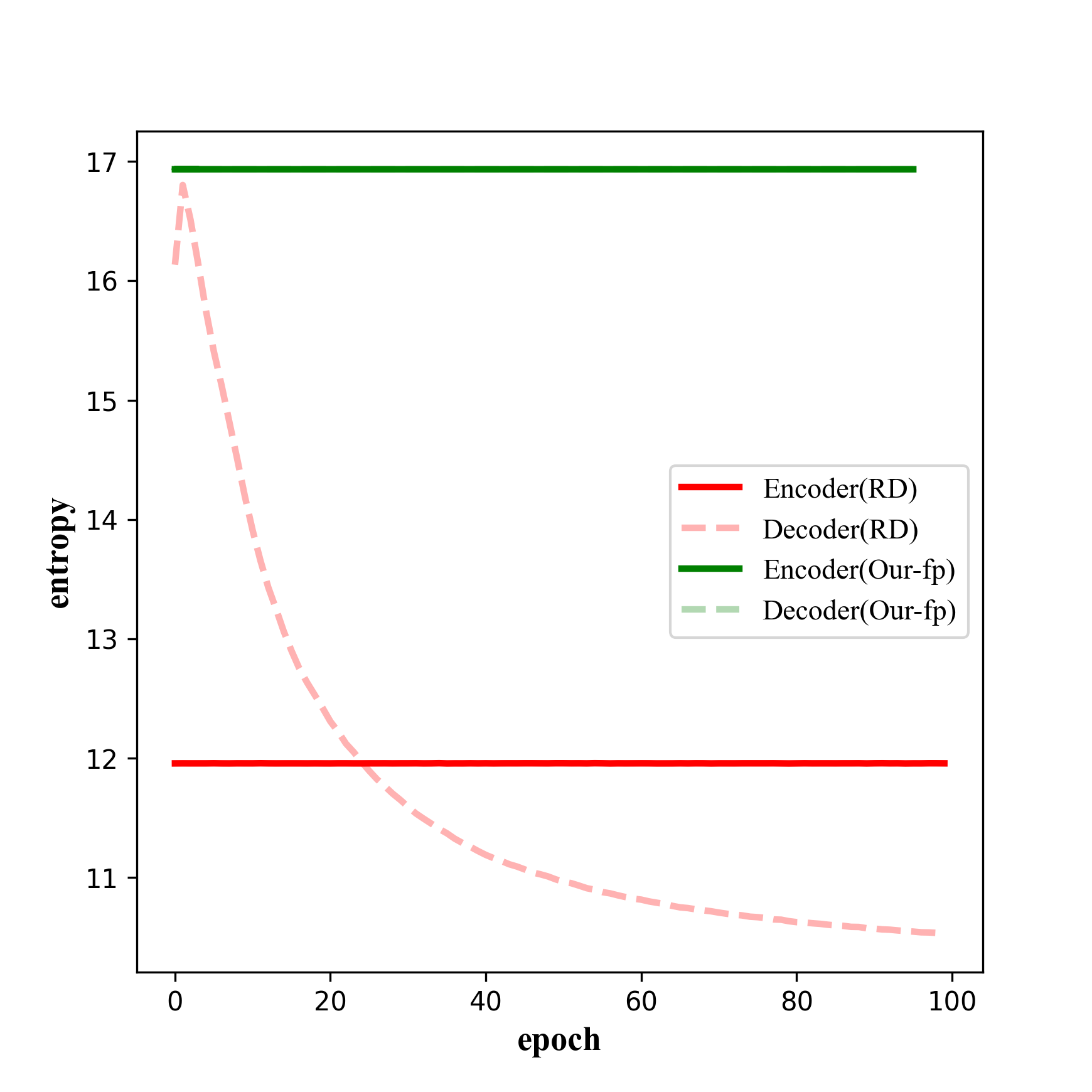}}
  \caption{Feature alignment process of RD and MiniMaxAD when trained on \textit{drink can}, viewed from the perspective of entropy, shows that the two green lines overlap. }
  \label{figa3}
\end{figure}

A high variance indicate greater diversity and potential for capturing complex patterns or relationships. More directly, information entropy, which focuses on the entropic properties of probability distributions (i.e., the uncertainty associated with these distributions), provides a more direct measure of the data's informational content. firstly, we calculate the occurrence $c_i$ of each feature $\mathcal{f}_i$ in the flattened feature maps $\mathcal{F}({f^k_E/\left\|f^k_E\right\|})$ and $\mathcal{F}({f^k_D/\left\|f^k_D\right\|})$, with a total count of $N$, where $N = C^k \times H^k \times W^k$. The probability of each feature $\mathcal{f}_i$ is then computed as:
\begin{equation}
  p(\mathcal{f}_i) = \frac{c_i}{N}
\end{equation}
Subsequently, the entropy of the feature map is given by:
\begin{equation}
  H(\mathcal{F}) = -\sum_{i=1}^{n} p(\mathcal{f}_i) \log_2 (p(\mathcal{f}_i) + \epsilon)
\end{equation}
where $\epsilon$ represents a very small positive number, n is the total number of different pixel values in feature maps. We calculated the mean of the entropy of the three layers of feature maps. The results depicted in Figure~\ref{figa3} demonstrate that, with the aid of the GRN unit, MiniMaxAD consistently generates feature maps that are richer in informative content. In contrast, the RD approach exhibits a trend towards degradation.

\subsection{Additional Stability Experiments}
The results in the main paper are reported with a single random seed following our baseline \cite{dengAnomalyDetectionReverse2022}. In Tables~\ref{taba1} and \ref{taba2}, we report the mean and standard deviation of three runs with three different random seeds (1, 11 and 111) \cite{guoReContrastDomainspecificAnomaly2023}.

\begin{table}[t]
  \centering
  \caption{ Performance on AeBAD-S over three runs. }
  \begin{tabular}{l|cc}
    \toprule
                 & I-AUROC            & AUPRO             \\
    \midrule
    Same         & 86.27  $\pm$  0.45 & 78.47 $\pm$  0.21 \\
    Background   & 89.03  $\pm$  0.41 & 79.43 $\pm$  0.42 \\
    Illumination & 85.87  $\pm$  0.68 & 83.73 $\pm$  0.29 \\
    View         & 86.93  $\pm$  0.54 & 76.27 $\pm$  0.31 \\
    \midrule
    Average      & 87.03  $\pm$  0.29 & 79.42 $\pm$  0.29 \\
    \bottomrule
  \end{tabular}
  \label{taba1}
\end{table}

\begin{table}[t]
  \centering
  \caption{ Performance on GoodsAD over three runs. }
  \begin{tabular}{l|cc}
    \toprule
                  & I-AUROC          & AUPRO             \\
    \midrule
    cigarette box & 97.77 $\pm$ 0.34 & 85.77 $\pm$  0.38 \\
    drink bottle  & 80.40 $\pm$ 0.33 & 86.80 $\pm$  0.08 \\
    drink can     & 89.80 $\pm$ 0.41 & 87.00 $\pm$  0.49 \\
    food bottle   & 91.20 $\pm$ 0.45 & 92.60 $\pm$  0.16 \\
    food box      & 78.67 $\pm$ 0.96 & 84.03 $\pm$  0.05 \\
    food package  & 76.50 $\pm$ 1.07 & 86.90 $\pm$  0.29 \\
    \midrule
    Average       & 85.71 $\pm$ 0.23 & 87.2 $\pm$  0.16  \\
    \bottomrule
  \end{tabular}
  \label{taba2}
\end{table}

\begin{table}[ht]
  \small
  \centering
  \caption{Ablation study on the values of $p_{lim}$ in $\mathcal{L}_{local-hm}$ on GoodsAD .}
  \begin{tabular}{@{}l|cccc@{}}
    \toprule
    $p_{lim}$ & 0.99 & 0.999 & 0.9995 & 0.9999 \\ \midrule
    I-AUROC   & 77.9 & 81.5  & 80.6   & 77.3   \\ \bottomrule
  \end{tabular}
  \label{taba5}
\end{table}

\subsection{Additional Ablation Study}
As shown in Table~\ref{taba4}, we performed an ablation study of $p_{hard}$ and $p_{lim}$ in $\mathcal{L}_{adc}$. Intuitively, $p_{hard}$ controls the decay rate and $p_{lim}$ represents the lower limit of decay. The results show that $\mathcal{L}_{adc}$ has the most stable performance at $p_{lim}=0.9995$,  where its performance is less affected by changes in \(p_{hard}\). In addition, our ablation experiments on $\mathcal{L}_{local-hm}$ in Table~\ref{taba5} show that fixed mining strategy yields unsatisfactory results.

\begin{table*}[!t]
  \small
  \centering
  \caption{Ablation on the values of $p_{hard}$ and $p_{lim}$ in $\mathcal{L}_{adc}$ on GoodsAD.}
  \begin{tabular}{@{}l|ccc|ccc|ccc@{}}
    \toprule
    $p_{lim}$  & \multicolumn{3}{c|}{0.999} & \multicolumn{3}{c|}{0.9995} & \multicolumn{3}{c}{0.9999}                                                                 \\ \midrule
    $p_{hard}$ & 0.9995                     & 0.9999                      & 1                          & 0.9995        & 0.9999        & 1    & 0.9995 & 0.9999 & 1    \\ \midrule
    I-AUROC    & 85.5                       & 85.9                        & 85.9                       & \textbf{86.1} & \textbf{86.1} & 85.8 & 85.7   & 85.4   & 85.3 \\ \bottomrule
  \end{tabular}
  \label{taba4}
\end{table*}

\section{More Quantitative Results}


\textbf{More quantitative results on the MVTec-AD dataset.} Table \ref{tab:mvtecsp} and Tables \ref{tab:mvtecpx1} and \ref{tab:mvtecpx2} present the image-level anomaly detection results and pixel-level anomaly localization quantitative results, respectively, for all categories in the MVTec-AD dataset under the unified multi-class setting.

\textbf{More quantitative results on the ViSA dataset.} Table \ref{tab:visasp} and Tables \ref{tab:visapx1} and \ref{tab:visapx2} present the image-level anomaly detection results and pixel-level anomaly localization quantitative results, respectively, for all categories in the ViSA dataset under the unified multi-class setting.

\textbf{More quantitative results on the Real-IAD dataset.} Table \ref{tab:realiadsp}, along with Tables \ref{tab:realiadpx1} and \ref{tab:realiadpx2}, presents the image-level anomaly detection results and pixel-level anomaly localization quantitative results for all categories of the Real-IAD dataset under the unified multi-class setting. Similarly, Table \ref{tab:s_realiadsp}, alongside Tables \ref{tab:s_realiadpx} and \ref{tab:s_realiadpx2}, provides the image-level anomaly detection results and pixel-level anomaly localization quantitative results for all categories of the Real-IAD dataset under the separated single-class setting.


\begin{table*}[h]
  \centering
  \caption{Comparison with SoTA methods on MVTec-AD dataset for multi-class anomaly detection with AUROC/AP/F1\_max metrics.}
  \resizebox{\linewidth}{!}{
    \begin{tabular}{ccccccc}
      \toprule
      Method$\rightarrow$  & RD                                                 & UniAD                                            & SimpleNet                                        & DiAD                                            & MambaAD                           & MMAD-fr                                            \\
      \cline{1-1}
      Category$\downarrow$ & CVPR'22                                            & NeurlPS'22                                       & CVPR'23                                          & AAAI'24                                         & NeurlPS'24                        & Ours                                               \\
      \hline
      Bottle               & 99.6            /99.9            /98.4             & 99.7            /\textbf{100.}/\textbf{100.}     & \textbf{100.}  /\textbf{100.}/\textbf{100.}      & 99.7            /96.5            /91.8          & \textbf{100.}  /\textbf{100.}/\textbf{100.}     & \textbf{100.}  /\textbf{100.}  /\textbf{100.}      \\
      Cable                & 84.1            /89.5            /82.5             & 95.2            /95.9          /88.0             & 97.5            /98.5          /94.7             & 94.8            /98.8            /95.2          & \underline{98.8}/\underline{99.2}/\underline{95.7} & \textbf{99.2}   /\textbf{99.6}   /\textbf{96.3}    \\
      Capsule              & 94.1            /96.9            /\textbf{96.9}    & 86.9            /97.8          /94.4             & 90.7            /97.9          /93.5             & 89.0            /97.5            /95.5          & \underline{94.4}/\underline{98.7}/94.9 & \textbf{96.7}   /\textbf{99.3}   /\underline{96.3} \\
\bottomrule
    \end{tabular}%
  }
  \label{tab:mvtecsp}
\end{table*}%

\begin{table*}[h]
  \centering
  \caption{Comparison with SoTA methods on MVTec-AD dataset for multi-class anomaly localization with AUROC/AP metrics.}
  \begin{tabular}{ccccccc}
    \toprule
    Method$\rightarrow$  & RD                                & UniAD                             & SimpleNet                      & DiAD                           & {MambaAD}                         & MMAD-fr                           \\
    \cline{1-1}
    Category$\downarrow$ & CVPR'22                           & NeurlPS'22                        & CVPR'23                        & AAAI'24                        & NeurlPS'24                        & Ours                              \\
    \hline
    Bottle               & 97.8            /68.2             & 98.1            /66.0             & 97.2            /53.8          & \underline{98.4}/52.2          & \textbf{98.8}   /\textbf{79.7}    & 97.8            /\underline{75.4} \\
    Cable                & 85.1            /26.3             & \textbf{97.3}   /39.9             & 96.7            /42.4          & \underline{96.8}/\textbf{50.1} & 95.8            /42.2             & 95.1            /\underline{48.4} \\
    Capsule              & \textbf{98.8}   /43.4             & \underline{98.5}/42.7             & \underline{98.5}/5.4           & 97.1            /42.0          & 98.4            /\underline{43.9} & 96.5            /\textbf{47.2}    \\
    Hazelnut             & 97.9            /36.2             & 98.1            /55.2             & 98.4            /44.6          & 98.3            /\textbf{79.2} & \textbf{99.0}   /63.6             & \underline{98.8}/\underline{69.0} \\
    MetalNut             & 94.8            /55.5             & 62.7            /14.6             & \textbf{98.0}   /\textbf{83.1} & \underline{97.3}/30.0          & 96.7            /\underline{74.5} & 91.9            /61.8             \\
    Pill                 & \textbf{97.5}   /63.4             & 95.0            /44.0             & 96.5            /\textbf{72.4} & 95.7            /46.0          & \underline{97.4}/64.0             & 96.1            /\underline{65.1} \\
\bottomrule
  \end{tabular}%
  \label{tab:mvtecpx1}%
\end{table*}%

\begin{table*}[h]
  \centering
  \caption{Comparison with SoTA methods on MVTec-AD dataset for multi-class anomaly localization with F1\_max/AUPRO metrics.}
  \begin{tabular}{ccccccc}
    \toprule
    Method$\rightarrow$  & RD                                & UniAD                          & SimpleNet                         & DiAD                              & MambaAD                           & MMAD-fr                           \\
    \cline{1-1}
    Category$\downarrow$ & CVPR'22                           & NeurlPS'22                     & CVPR'23                           & AAAI'24                           & NeurlPS'24                        & Ours                              \\
    \hline
    Bottle               & 67.6            /94.0             & 69.2            /93.1          & 62.4            /89.0             & 54.8            /86.6             & \textbf{76.7}   /\textbf{95.2}    & \underline{73.2}/\underline{95.0} \\
    Cable                & 33.6            /75.1             & 45.2            /86.1          & \underline{51.2}/85.4             & \textbf{57.8}   /80.5             & 48.1            /\textbf{90.3}    & 50.9            /\underline{86.3} \\
    Capsule              & \underline{50.0}/\textbf{94.8}    & 46.5            /92.1          & 44.3            /84.5             & 45.3            /87.2             & 47.7            /92.6             & \textbf{52.3}   /\underline{93.2} \\
    Hazelnut             & 51.6            /92.7             & 56.8            /94.1          & 51.4            /87.4             & \textbf{80.4}   /91.5             & 64.4            /\underline{95.7} & \underline{66.2}/\textbf{96.6}    \\
    MetalNut             & 66.4            /\underline{91.9} & 29.2            /81.8          & \textbf{79.4}   /85.2             & 38.3            /90.6             & \underline{79.1}/\textbf{93.7}    & 61.4            /91.5             \\
    Pill                 & 65.2            /\textbf{95.8}    & 53.9            /95.3          & \textbf{67.7}   /81.9             & 51.4            /89.0             & \underline{66.5}/\underline{95.7} & 66.1            /94.2             \\
\bottomrule
  \end{tabular}%
  \label{tab:mvtecpx2}%
\end{table*}%

\begin{table*}[h]
  \centering
  \caption{Comparison with SoTA methods on {VisA} dataset for multi-class anomaly detection with AUROC/AP/F1\_max metrics.}
  \resizebox{\linewidth}{!}{
    \begin{tabular}{ccccccc}
      \toprule
      Method$\rightarrow$  & RD                                                  & UniAD                      & SimpleNet      & DiAD                                            & MambaAD                                            & MMAD-fr                                            \\
      \cline{1-1}
      Category$\downarrow$ & CVPR'22                                             & NeurlPS'22                 & CVPR'23        & AAAI'24                                         & NeurlPS'24                                         & Ours                                               \\
      \hline
      pcb1                 & \underline{96.2} /\underline{95.5}/\underline{91.9} & 92.8/92.7/87.8             & 91.6/91.9/86.0 & 88.1         /88.7            /80.7             & 95.4            /93.0            /91.6             & \textbf{98.0}   /\textbf{97.8}   /\textbf{95.1}    \\
      pcb2                 & \textbf{97.8}    /\textbf{97.8}   /\textbf{94.2}    & 87.8/87.7/83.1             & 92.4/93.3/84.5 & 91.4         /91.4            /84.7             & 94.2            /93.7            /89.3             & \underline{97.1}/\underline{97.1}/\underline{93.0} \\
      pcb3                 & \underline{96.4} /\underline{96.2}/\underline{91.0} & 78.6/78.6/76.1             & 89.1/91.1/82.6 & 86.2         /87.6            /77.6             & 93.7            /94.1            /86.7             & \textbf{97.8}   /\textbf{97.7}   /\textbf{94.5}    \\
      pcb4                 & \textbf{99.9}    /\textbf{99.9}   /\textbf{99.0}    & 98.8/98.8/94.3             & 97.0/97.0/93.5 & 99.6         /99.5            /97.0             & \textbf{99.9}   /\textbf{99.9}   /\underline{98.5} & 99.8            /99.8            /\underline{98.5} \\
\bottomrule
    \end{tabular}%
  }
  \label{tab:visasp}%
\end{table*}

\begin{table*}[h]
  \centering
  \caption{Comparison with SoTA methods on VisA dataset for multi-class anomaly localization with AUROC/AP metrics.}
  \begin{tabular}{ccccccc}
    \toprule
    Method$\rightarrow$  & RD                                & UniAD                 & SimpleNet                      & DiAD                           & {MambaAD}                         & MMAD-fr                           \\
    \cline{1-1}
    Category$\downarrow$ & CVPR'22                           & NeurlPS'22            & CVPR'23                        & AAAI'24                        & NeurlPS'24                        & Ours                              \\
    \hline
    pcb1                 & \underline{99.4}/66.2             & 93.3            /3.9  & 99.2         /\textbf{86.1}    & 98.7         /49.6             & \textbf{99.8}   /77.1             & 99.0            /\underline{79.0} \\
    pcb2                 & \underline{98.0}/\underline{22.3} & 93.9            /4.2  & 96.6         /8.9              & 95.2         /7.5              & \textbf{98.9}   /13.3             & 97.1            /\textbf{36.6}    \\
    pcb3                 & 97.9            /26.2             & 97.3            /13.8 & 97.2         /\underline{31.0} & 96.7         /8.0              & \textbf{99.1}   /18.3             & \underline{98.9}/\textbf{39.6}    \\
    pcb4                 & 97.8            /31.4             & 94.9            /14.7 & 93.9         /23.9             & 97.0         /17.6             & \textbf{98.6}   /\underline{47.0} & \underline{98.2}/\textbf{54.5}    \\
    macaroni1            & \underline{99.4}/2.9              & 97.4            /3.7  & 98.9         /3.5              & 94.1         /10.2             & \textbf{99.5}   /\underline{17.5} & \underline{99.4}/\textbf{42.5}    \\
    macaroni2            & \textbf{99.7}   /\underline{13.2} & 95.2            /0.9  & 93.2         /0.6              & 93.6         /0.9              & \underline{99.5}/9.2              & 98.9            /\textbf{29.1}    \\
\bottomrule
  \end{tabular}%
  \label{tab:visapx1}%
\end{table*}%

\begin{table*}[h]
  \centering
  \caption{Comparison with SoTA methods on VisA dataset for multi-class anomaly localization with F1\_max/AUPRO metrics.}
  \begin{tabular}{ccccccc}
    \toprule
    Method$\rightarrow$  & RD                                & UniAD              & SimpleNet             & DiAD                  & {MambaAD}                         & MMAD-fr                           \\
    \cline{1-1}
    Category$\downarrow$ & CVPR'22                           & NeurlPS'22         & CVPR'23               & AAAI'24               & NeurlPS'24                        & Ours                              \\
    \hline
    pcb1                 & 62.4            /\textbf{95.8}    & 8.3 /64.1          & \textbf{78.8}   /83.6 & 52.8            /80.2 & 72.4           /\underline{92.8}  & \underline{73.3}/87.8             \\
    pcb2                 & \underline{30.0}/\textbf{90.8}    & 9.2 /66.9          & 18.6            /85.7 & 16.7            /67.0 & 23.4           /\underline{89.6}  & \textbf{41.6}   /82.8             \\
    pcb3                 & 35.2            /\textbf{93.9}    & 21.9/70.6          & \underline{36.1}/85.1 & 18.8            /68.9 & 27.4           /89.1              & \textbf{40.0}   /\underline{93.0} \\
    pcb4                 & 37.0            /\underline{88.7} & 22.9/72.3          & 32.9            /61.1 & 27.2            /85.0 & \underline{46.9}/87.6             & \textbf{54.8}   /\textbf{91.8}    \\
    macaroni1            & 6.9             /\underline{95.3} & 9.7 /84.0          & 8.4             /92.0 & 16.7            /68.5 & \underline{27.6}/95.2             & \textbf{45.4}   /\textbf{95.4}    \\
    macaroni2            & \underline{21.8}/\textbf{97.4}    & 4.3 /76.6          & 3.9             /77.8 & 2.8             /73.1 & 16.1           /96.2              & \textbf{38.7}   /\underline{97.0} \\
\bottomrule
  \end{tabular}%
  \label{tab:visapx2}%
\end{table*}%

\begin{table*}[h]
  \centering
  \caption{Comparison with SoTA methods on Real-IAD dataset for multi-class anomaly detection with AUROC/AP/F1\_max metrics.}
  \resizebox{\linewidth}{!}{
    \begin{tabular}{ccccccc}
      \toprule
      Method$\rightarrow$  & RD4AD                                              & UniAD                                              & SimpleNet      & DiAD                                            & {MambaAD}                                          & MMAD-fr                                            \\
      \cline{1-1}
      Category$\downarrow$ & CVPR'22                                            & NeurlPS'22                                         & CVPR'23        & AAAI'24                                         & NeurlPS'24                                         & Ours                                               \\
      \hline
      audiojack            & 76.2            /63.2            /60.8             & 81.4            /\underline{76.6}/64.9             & 58.4/44.2/50.9 & 76.5            /54.3         /65.7             & \underline{84.2}/76.5            /\underline{67.4} & \textbf{86.5}   /\textbf{82.8}   /\textbf{71.1}    \\
      bottle cap           & 89.5            /86.3            /81.0             & \underline{92.5}/91.7            /81.7             & 54.1/47.6/60.3 & 91.6            /\textbf{94.0}/\textbf{87.9}    & \textbf{92.8}   /\underline{92.0}/\underline{82.1} & 87.5            /85.9            /78.7             \\
      button battery       & 73.3            /78.9            /76.1             & 75.9            /81.6            /\underline{76.3} & 52.5/60.5/72.4 & \textbf{80.5}   /71.3         /70.6             & \underline{79.8}/\textbf{85.3}   /\textbf{77.8}    & 76.8            /\underline{83.6}/75.8             \\
      end cap              & 79.8            /\underline{84.0}/77.8             & \underline{80.9}/\textbf{86.1}   /\underline{78.0} & 51.6/60.8/72.9 & \textbf{85.1}   /83.4         /\textbf{84.8}    & 78.0            /82.8            /77.2             & 78.1            /83.3            /76.4             \\
\bottomrule
    \end{tabular}%
  }
  \label{tab:realiadsp}%
\end{table*}%

\begin{table*}[h]
  \centering
  \caption{Comparison with SoTA methods on Real-IAD dataset for multi-class anomaly localization with AUROC/AP metrics.}
  \begin{tabular}{ccccccc}
    \toprule
    Method$\rightarrow$  & RD4AD                             & UniAD                 & SimpleNet & DiAD                              & MambaAD                           & MMAD-fr                           \\
    \cline{1-1}
    Category$\downarrow$ & CVPR'22                           & NeurlPS'22            & CVPR'23   & AAAI'24                           & NeurlPS'24                        & Ours                              \\
    \hline
    audiojack            & 96.6            /12.8             & 97.6            /20.0 & 74.4/0.9  & 95.5            /\underline{25.4} & \underline{97.7}/21.6             & \textbf{98.2}   /\textbf{41.4}    \\
    bottle cap           & \underline{99.5}/18.9             & \underline{99.5}/19.4 & 85.3/2.3  & 94.5            /25.3             & \textbf{99.7}   /\textbf{30.6}    & 99.1            /\underline{30.1} \\
    button battery       & 97.6            /33.8             & 96.7            /28.5 & 75.9/3.2  & \textbf{98.3}   /\textbf{63.9}    & \underline{98.1}/46.7             & 97.3            /\underline{52.8} \\
    end cap              & \underline{96.7}/12.5             & 95.8            /8.8  & 63.1/0.5  & 89.6            /\underline{14.4} & \textbf{97.0}   /12.0             & 95.1            /\textbf{19.0}    \\
    eraser               & \textbf{99.5}   /30.8             & \underline{99.3}/24.4 & 80.6/2.7  & 95.8            /\textbf{52.7}    & 99.2            /30.2             & 98.4            /\underline{40.9} \\
    fire hood            & \textbf{98.9}   /\underline{27.7} & 98.6            /23.4 & 70.5/0.3  & 97.3            /27.1             & \underline{98.7}/25.1             & 98.3            /\textbf{35.4}    \\
\bottomrule
  \end{tabular}%
  \label{tab:realiadpx1}%
\end{table*}%

\begin{table*}[h]
  \centering
  \caption{Comparison with SoTA methods on Real-IAD dataset for multi-class anomaly localization with F1\_max/AUPRO metrics.}
  \begin{tabular}{ccccccc}
    \toprule
    Method$\rightarrow$  & RD4AD                 & UniAD                 & SimpleNet & DiAD                  & MambaAD                           & MMAD-fr                            \\
    \cline{1-1}
    Category$\downarrow$ & CVPR'22               & NeurlPS'22            & CVPR'23   & AAAI'24               & NeurlPS'24                        & Ours                               \\
    \hline
    audiojack            & 22.1/79.6             & 31.0/83.7             & 4.8 /38.0 & \underline{31.9}/52.6 & 29.5            /\underline{83.9} & \textbf{47.1}    /\textbf{89.4}    \\
    bottle cap           & 29.9/95.7             & 29.6/\underline{96.0} & 5.7 /45.1 & 31.1            /25.3 & \textbf{34.6}   /\textbf{97.2}    & \underline{34.2} /94.9             \\
    button battery       & 37.8/\underline{86.5} & 34.4/77.5             & 6.6 /40.5 & \textbf{60.4}   /36.9 & 49.5            /86.2             & \underline{53.7} /\textbf{88.2}    \\
    end cap              & 22.5/\underline{89.2} & 17.4/85.4             & 2.8 /25.7 & \underline{22.7}/29.5 & 19.6            /\textbf{89.4}    & \textbf{27.5}    /86.1             \\
    eraser               & 36.7/\textbf{96.0}    & 30.9/\underline{94.1} & 7.1 /42.8 & \textbf{53.9}   /46.7 & 38.3            /93.7             & \underline{45.8} /90.6             \\
    fire hood            & 35.2/\underline{87.9} & 32.2/85.3             & 2.2 /25.3 & \underline{35.3}/34.7 & 31.3            /86.3             & \textbf{41.8}    /\textbf{90.5}    \\
\bottomrule
  \end{tabular}%
  \label{tab:realiadpx2}%
\end{table*}%

\begin{table*}[h]
  \centering
  \caption{Comparison with SoTA methods on Real-IAD dataset for single-class anomaly detection with AUROC/AP/F1\_max metrics.}
  \resizebox{\linewidth}{!}{
    \begin{tabular}{ccccccc}
      \toprule
      Method$\rightarrow$  & RD4AD                                              & UniAD          & SimpleNet                                          & PatchCore                                          & MVAD                                               & MMAD-fr                                            \\
      \cline{1-1}
      Category$\downarrow$ & CVPR'22                                            & NeurlPS'22     & CVPR'23                                            & CVPR'22                                            & Arxiv'24                                         & Ours                                               \\
      \hline
      audiojack            & 83.4            /78.0            /66.6             & 78.7/60.8/64.0 & \underline{88.4}/84.4            /74.7             & 87.6            /\underline{85.0}/\underline{75.2} & 86.9            /82.5            /70.6             & \textbf{92.1}   /\textbf{89.3}   /\textbf{80.4}    \\
      bottle cap           & \textbf{96.3}   /\textbf{96.3}   /\textbf{89.0}    & 85.6/82.6/74.9 & 91.1            /88.6            /83.0             & 95.3            /95.0            /86.2             & \underline{95.6}/\underline{95.4}/\underline{87.0} & 94.5            /94.2            /85.7             \\
      button battery       & 88.6            /91.3            /82.9             & 65.9/71.9/74.6 & 88.4            /89.9            /83.5             & 88.5            /90.8            /84.0             & \textbf{90.8}   /\underline{92.1}/\textbf{86.8}    & \textbf{90.8}   /\textbf{92.6}   /\underline{85.8} \\
      end cap              & 82.9            /86.9            /80.0             & 80.6/84.4/78.3 & 83.7            /88.4            /79.5             & \textbf{86.9}   /\textbf{90.0}   /\underline{82.3} & 85.8            /88.8            /81.8             & \underline{86.6}/\underline{89.2}/\textbf{82.5}    \\
\bottomrule
    \end{tabular}%
  }
  \label{tab:s_realiadsp}%
\end{table*}%

\begin{table*}[h]
  \centering
  \caption{Comparison with SoTA methods on Real-IAD dataset for single-class anomaly localization with AUROC/AP metrics.}
  \begin{tabular}{ccccccc}
    \toprule
    Method$\rightarrow$  & RD4AD                             & UniAD                 & SimpleNet             & PatchCore                         & MVAD                           & MMAD-fr                           \\
    \cline{1-1}
    Category$\downarrow$ & CVPR'22                           & NeurlPS'22            & CVPR'23               & CVPR'22                           & Arxiv'24                     & Ours                              \\
    \hline
    audiojack            & 98.2            /40.7             & 97.2            /7.8  & 98.2            /19.7 & 98.0            /\underline{41.0} & \underline{98.8}/36.6          & \textbf{99.3}   /\textbf{50.2}    \\
    bottle cap           & \textbf{99.7}   /\textbf{41.9}    & 99.2            /21.4 & 98.5            /14.7 & 99.2            /28.1             & \textbf{99.7}   /36.2          & 99.2            /\underline{41.8} \\
    button battery       & \underline{98.7}/52.6             & 93.7            /13.6 & 98.0            /21.3 & 98.4            /\underline{53.2} & \textbf{98.8}   /46.6          & \underline{98.7}/\textbf{60.3}    \\
    end cap              & 97.8            /\textbf{23.4}    & 96.7            /7.6  & 94.2            /7.2  & 97.0            /15.9             & \underline{98.1}/12.7          & \textbf{98.3}   /\underline{22.4} \\
    eraser               & \underline{99.2}/\underline{40.6} & 99.0            /17.7 & 98.3            /16.5 & \textbf{99.3}   /36.2             & \underline{99.2}/30.7          & 98.8            /\textbf{42.0}    \\
    fire hood            & 98.7            /\underline{32.6} & 98.5            /22.7 & 97.5            /10.2 & 97.8            /26.6             & \textbf{99.1}   /28.5          & \underline{98.9}/\textbf{44.2}    \\
\bottomrule
  \end{tabular}%
  \label{tab:s_realiadpx}%
\end{table*}%

\begin{table*}[h]
  \centering
  \caption{Comparison with SoTA methods on Real-IAD dataset for single-class anomaly localization with F1\_max/AUPRO metrics.}
  \begin{tabular}{ccccccc}
    \toprule
    Method$\rightarrow$  & RD4AD                              & UniAD                  & SimpleNet             & PatchCore                         & MVAD                              & MMAD-fr                           \\
    \cline{1-1}
    Category$\downarrow$ & CVPR'22                            & NeurlPS'22             & CVPR'23               & CVPR'22                           & Arxiv'24                        & Ours                              \\
    \hline
    audiojack            & \underline{47.7} /89.5             & 14.4 /83.9             & 28.9/86.9             & 46.0            /86.7             & 44.4            /\underline{90.6} & \textbf{51.3}   /\textbf{95.0}    \\
    bottle cap           & \underline{40.5} /\textbf{98.2}    & 30.9 /93.8             & 25.7/85.8             & 31.5            /95.3             & 39.5            /\underline{97.4} & \textbf{41.3}   /96.0             \\
    button battery       & 51.5             /\textbf{93.2}    & 20.7 /70.3             & 37.3/72.5             & \underline{52.1}/86.0             & 45.7            /88.5             & \textbf{58.2}   /\underline{92.3} \\
    end cap              & \textbf{31.5}    /\underline{93.2} & 15.2 /88.3             & 15.3/80.5             & 24.1            /90.1             & 21.5            /92.6             & \underline{29.2}/\textbf{94.5}    \\
    eraser               & \textbf{46.3}    /\underline{94.3} & 24.6 /92.5             & 22.0/88.0             & 41.8            /\textbf{94.4}    & 35.9            /92.3             & \underline{45.8}/93.8             \\
    fire hood            & \underline{41.8} /\underline{92.7} & 31.2 /84.6             & 16.7/77.3             & 34.5            /85.2             & 37.1            /90.1             & \textbf{48.1}   /\textbf{93.0}    \\
\bottomrule
  \end{tabular}%

  \label{tab:s_realiadpx2}%
\end{table*}%


\end{document}